\documentclass[10pt,journal,compsoc]{IEEEtran}
\ifCLASSOPTIONcompsoc
  \usepackage[nocompress]{cite}
\else
  \usepackage{cite}
\fi
\ifCLASSINFOpdf
   \usepackage[pdftex]{graphicx}
   \graphicspath{{../pdf/}{../jpeg/}}
\else
   \usepackage[dvips]{graphicx}
   \graphicspath{{../eps/}}
\fi
\usepackage{amsmath}
\usepackage{amssymb}
\usepackage{amsfonts}
\usepackage{fdsymbol}

\usepackage{algorithmic}
\usepackage{algorithm}
\makeatletter
\newcommand{\removelatexerror}{\let\@latex@error\@gobble}
\makeatother

\usepackage{array}

\ifCLASSOPTIONcompsoc
  \usepackage[caption=false,font=footnotesize,labelfont=sf,textfont=sf]{subfig}
\else
  \usepackage[caption=false,font=footnotesize]{subfig}
\fi
\usepackage{caption}

\usepackage{color}

\usepackage{url}

\newtheorem{definition}{Definition}[section]
\newtheorem{proposition}{Proposition}[section]
\newtheorem{lemma}{Lemma}[section]
\newtheorem{theorem}{Theorem}[section]
\newtheorem{assumption}{Assumption}[section]

\begin{document}
%
\title{Robust Local Preserving and Global Aligning Network for Adversarial Domain Adaptation}
%
%
%
%

\author{Wenwen~Qiang$^*$,
        Jiangmeng~Li$^*$,
        Changwen~Zheng,
        Bing~Su,
        and~Hui~Xiong,~\IEEEmembership{Fellow,~IEEE}
\IEEEcompsocitemizethanks{\IEEEcompsocthanksitem W. Qiang and J. Li are with the University of Chinese Academy of Sciences, Beijing, China. They are also with the Science \& Technology on Integrated Information System Laboratory, Institute of Software Chinese Academy of Sciences, Beijing, China. E-mail: a01114115@163.com, Jiangmeng2019@iscas.ac.cn
\IEEEcompsocthanksitem C. Zheng is with the Science \& Technology on Integrated Information System Laboratory, Institute of Software Chinese Academy of Sciences, Beijing, China. E-mail: changwen@iscas.ac.cn
\IEEEcompsocthanksitem B. Su is with the Beijing Key Laboratory of Big Data Management and Analysis Methods, Gaoling School of Artificial Intelligence, Renmin University of China, Beijing, 100872, China. E-mail: subingats@gmail.com
\IEEEcompsocthanksitem H. Xiong is with the Hong Kong University of Science and Technology (Guangzhou). E-mail: xionghui@ust.hk.
\IEEEcompsocthanksitem $^*$They have contributed equally to this work (Corresponding author: Bing Su).
\IEEEcompsocthanksitem \copyright 2021 IEEE. Personal use of this material is permitted. Permission from IEEE must be obtained for all other uses, in any current or future media, including reprinting/republishing this material for advertising or promotional purposes, creating new collective works, for resale or redistribution to servers or lists, or reuse of any copyrighted component of this work in other works.}}

%
%

\markboth{SUBMITTED TO IEEE TRANSACTIONS ON KNOWLEDGE AND DATA ENGINEERING}%
{Shell \MakeLowercase{\textit{et al.}}: Bare Demo of IEEEtran.cls for Computer Society Journals}
%



\IEEEtitleabstractindextext{%
\begin{abstract}
Unsupervised domain adaptation (UDA) requires source domain samples with clean ground truth labels during training. Accurately labeling a large number of source domain samples is time-consuming and laborious. An alternative is to utilize samples with noisy labels for training. However, training with noisy labels can greatly reduce the performance of UDA. In this paper, we address the problem that learning UDA models only with access to noisy labels and propose a novel method called robust local preserving and global aligning network (RLPGA). RLPGA improves the robustness of the label noise from two aspects. One is learning a classifier by a robust informative-theoretic-based loss function. The other is constructing two adjacency weight matrices and two negative weight matrices by the proposed local preserving module to preserve the local topology structures of input data. We conduct theoretical analysis on the robustness of the proposed RLPGA and prove that the robust informative-theoretic-based loss and the local preserving module are beneficial to reduce the empirical risk of the target domain. A series of empirical studies show the effectiveness of our proposed RLPGA.
\end{abstract}

}

\maketitle

\IEEEdisplaynontitleabstractindextext

%
\IEEEpeerreviewmaketitle

\IEEEraisesectionheading{\section{Introduction}\label{sec:introduction}}

\IEEEPARstart{U}{nsupervised} domain adaptation emphasizes \cite{ben01, pan2009survey, zhuang2020comprehensive} the problem of learning a classifier that can be transferred across two domains. In general, the samples in the source domain are labeled, while the samples in the target domain are unlabeled. The main challenge in this research area is to reduce the difference between the probability distributions of two domains \cite{ben2010, Nia1, shi2, Wang18}. To this end, the strategy based on discrepancy minimization has attracted much attention. Among them, adversarial learning methods have achieved remarkable performance improvements \cite{shen09}.

The training set of unsupervised adversarial domain adaptation models consists of two parts including the labeled source domain samples and unlabeled target domain samples. However, it is usually very expensive and tedious to accurately label large source domain training samples. An alternative way is to collect labels of samples from some crowdsourcing platforms in which the cost is cheaper and easier but the obtained labels are always contaminated by noise. As a result, the performance of adversarial domain adaptation models learning from noisy labels will be decreased. One reason is that adversarial domain adaptation models usually learn the classifier by minimizing the cross-entropy loss. The cross-entropy loss can be regarded as the distance between the outputs of the classifier and the labels, so it is sensitive to label noises. That is to say, when a careless annotator tends to label positive class to negative class, then the distance-based loss would force adversarial domain adaptation to learn a classifier who is more likely to output negative class than to output true label.

However, to make the domain adaptive models robust to label noises, it is not enough to only employ robust classification loss to train the models. The main reason is that robust loss can only reduce the impact of noisy labels, but can not completely eliminate it. Actually, for the source domain, learning with noisy labels can reduce the feature discriminability of samples in the latent space. This can lead to that a sample belonging to the one class is easy to be misclassified into another class. From a geometrically intuitive point of view, if a sample belonging to class $i$ is incorrectly labeled as class $j$, then the gradient backpropagation operation will force the sample to go from a place surrounded by many samples of the same type to a place surrounded by many different class samples. Therefore, besides learning based on a robust classification loss, designing an unsupervised method to maintain the local structure of the data distribution is also very important.

To tackle these issues, we propose a novel method for adversarial domain adaptation named \textit{Robust Local Preserving and Global Aligning Network} (RLPGA). RLPGA consists of three parts including a robust loss function for learning a classifier, a local preserving module, and a global aligning module. Firstly, RLPGA projects samples of both domains into a latent space. Then, a robust informative-theoretic-based loss function is minimized to learn the classifier. The global aligning module minimizes the Wasserstein distance between source domain distribution and target domain distribution. The local preserving module constructs two adjacency weight matrices and two negative weight matrices to encode the local topological relationship among samples and propose an objective function based on the graphs for local preserving. The major contributions of this paper are three-fold:
\begin{itemize}
	\item A robust informative-theoretic-based loss function is proposed to measure the performance of the classifier for adversarial domain adaptation.
	\item To reduce the effect of label noises from the perspective of the learned feature representation, we propose a new objective function for preserving local neighbor topology based on constructing two adjacency weight matrices and two negative weight matrices. We jointly minimize the Wasserstein distance between two domain distributions and the new objective. In this way, the margins between different classes are enlarged and hence the learned features are more discriminative and robust.
	\item We provide theoretical analysis on the robustness of RLPGA, and prove that the robust loss and the enhanced feature discriminability are beneficial to reduce the empirical risk in the target domain.
\end{itemize}

\section{Related works}
\textbf{Domain Adaptation Algorithm}. Existing domain adaptation methods can be divided into three categories. The first is instance-based methods \cite{chen02, chu03, huang01, tzeng04}, which enhance feature transferability by reweighting or subsampling the source domain samples. The second is parameter-based methods \cite{az01, chen2019, duan01, ro02, you2019}, which enhance the feature transferability by regularized terms or reweighting techniques. The last is representation learning based methods \cite{courty04, gani06, tzeng07, wu10, zhao09}, which first learn a latent space, and then align feature distributions across domains based on the learned feature representation by two strategies. 

For representation learning based methods, The first strategy to align feature distributions across domains is that moment matching based on statistical characteristics \cite{long2015, long2017}. \textit{E.g.}, Maximum mean discrepancy (MMD) \cite{gre01, tz02} measures the divergence of two distributions in the reproducing kernel Hilbert space (RKHS) with the advantages that it can approximate any moment of the distribution by choosing a suitable kernel function. Deep correlation alignment (DCORAL) \cite{sun201, sun2016} aligns two distributions by minimizing the difference in the second-order statistics of the two distributions. The other strategy is adversarial learning based on a zero-sum two-player game \cite{good01}. These adversarial methods have achieved remarkable performance improvements. The metric for adversarial learning based methods can be KL-divergence, H-divergence, and Wasserstein distance \cite{ben01, courty04, gani06, kuroki2019, shen09, long2017deep, sankn2018generate, lee2019sliced}. Among them, Wasserstein distance takes advantage of gradient superiority. \textit{E.g.}, Wasserstein Distance Guided Representation Learning (WDGRL) \cite{shen09} aligns the two distributions by minimizing the Wasserstein distance, which takes the advantage of gradient superiority. Sliced Wasserstein Discrepancy (SWD) \cite{lee2019sliced} is a method that based on sliced Wasserstein distance. As for domain adaptation over noisy labels, there are few pioneering works \cite{han2020towards, shu2019transferable} to handle this problem, and most of them focus on a robust classifier loss. \cite{ge2020mutual} proposes to tackle the label noise problem by clustering-based UDA methods for person re-ID. RLPGA handles robustness by considering both a robust classification loss and an enhanced feature discriminability. 

\textbf{Domain Adaptation Theory}. There are rich advances in domain adaptation theory. A rigorous error bound for unsupervised domain adaptation is proposed by \cite{ben201000, ma09}. Then, many extensions based on these bounds, from loss functions to Bayesian settings and regression problems, are put forward \cite{az01, cortes2015, ger2013, mo2012, zhang01}. \textit{E.g.}, Kuroki \cite{kuroki2019unsupervised} proposes a discrepancy measure called S-disc, which can not only provide a tighter generalization error bound but also have a convergence guarantee. Germain \cite{ger2013} proposed a PAC-Bayesian theory based on the domain disagreement pseudometric. Another related work is about the Wasserstein distance based domain adaptation algorithm and proves that Wasserstein distance can guarantee generalization for domain adaptation \cite{shen09}. As for our proposed method, the theoretical analysis for Wasserstein distance based domain adaptation can be directly extended to RLPGA. Also, RLPGA gives some theoretical analysis for robustness and advantage of enhancing the feature discriminability.

\begin{figure*}[tbh]	
	\begin{center}
		\centerline{\includegraphics[width=0.75\textwidth]{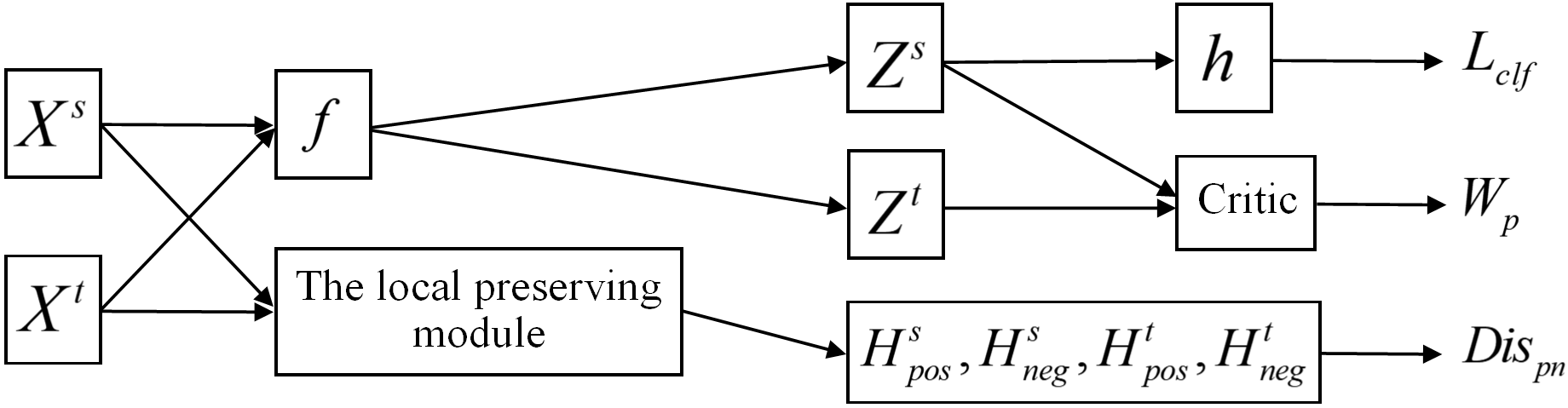}}
		\caption{The framework of RLPGA. $X^s$ and $X^t$ are the input datasets, $f$ is the feature extractor, $Z^s$ and $Z^t$ are the feature representations, $h$ is the classifier. First, RLPGA constructs four weight matrices by the local preserving module and maps the input samples to the latent space through the feature extractor. Then, RLPGA feeds $Z^s$ into $h$ to calculate the classification loss $L_{clf}$, feeds $Z^s$ and $Z^t$ into the Critic to calculate the Wasserstein distance between two domain distributions, and calculate the loss $Dis_{pn}$ based on the four weight matrices. Then, RLPGA minimizes the equation \ref{Eq:hh} to update the parameters of  $f$, $h$, and the Critic.}
		\label{fig:fram}		
	\end{center}	
	\vskip -0.35in
\end{figure*}

\section{Preliminaries}
\label{Sec:Preliminaries}

\subsection{Problem definition}
This paper considers the classification task of unsupervised domain adaptation (UDA). Let $\mathcal {X}$ represent the input feature space and $\eta :X \to Y$ be the domain-invariant ground truth labeling function, where $X \in \mathcal {X}$, and $Y$ is the label. Let ${P_s}$ be the input distribution over $X$ for the source domain and $P_t$ be the input distribution over $X$ for the target domain. Let $\mathcal {Z}$ be a latent space and $F: X \to Z$ be a class of feature extractors, where $Z \in \mathcal {Z}$. For a domain $u \in \left\{ {s,t} \right\}$, $P_u^{f\left( X \right)}  \left( Z \right) = P_u\left( {{f ^{ - 1}}\left( Z \right)} \right)$ represents the induced probability distribution over $\mathcal {Z}$, where $f  \in F $. For a given $Z \in \mathcal {Z}$. Denote $H :Z \to Y$ as a class of prediction functions. Then, the learned classifier can be represented as $h \left( {f \left( X \right)} \right)$, where $h \in H$. The goal is to learn a classifier that can minimize the following expected target risk:

\begin{equation}
\label{Eq:risk}
{R_{{P_t}}}\left( f,h \right) = \int {{P_t}\left( X \right)} \left| {\eta \left( X \right) - h \left( {f \left( X \right)} \right)} \right|dX.
\end{equation}
The difference between the supervised domain adaption (SDA) and UDA is that the label and the feature of the target domain dataset are all available during training phase for SDA, but for the UDA, we can only access to the feature of the target domain dataset during the training phase. For practical application, Dou \cite{dou2018unsupervised} propose to utilize the adversarial learning to UDA from the source Medical Image Analysis (MIA) domain to the target Computed Tomography (CT) domain.




Generally, UDA models learn a classifier $h $ using source domain samples with their ground truth labels. For real-world applications, it is costly to obtain the true data labels. Instead, we focus on using noisy labels to learn the classifier. We denote the source domain data as $X^s$ and its corresponding labels as $Y^{s*}$, where $Y^{s*}$ is the noisy version of the clean ground truth labels $Y^s$. Therefore, our goal is to learn a robust classifier based on the source domain data with noisy labels, which can be transferred to the target domain.

Formally, we denote ${Y^*}$ as the noisy version of $Y$ and ${P_{Y{Y^*}}}$ as the transition distribution from $Y$ to ${Y^*}$. Let $Y,{Y^*} \in \left\{ {1,2,...,C} \right\}$, and $C$ be the total number of the categories. Then we denote
$\Pr \left( {{Y^*} = {j}\left| {Y = i} \right.} \right) $ as the transition distribution to transfer the ground truth label $i$ to the class $j$, where $i,j \in \left\{ {1,2,...,C} \right\}$. We use ${{\rm{T}}_{Y \to {Y^*}}}$ to represent the $C \times C$ matrix, and ${{\rm{T}}_{Y \to {Y^*}}}\left( {i,j} \right) = \Pr \left( {{Y^*} = {j}\left| {Y = i} \right.} \right)$. Generally, the label noise can be defined into several kinds based on ${{\rm{T}}_{Y \to {Y^*}}}$ \cite{zhang20180}, \textit{e.g.}, define the label noise as class independent (or uniform), then ${{\rm{T}}_{Y \to {Y^*}}}$ can be written as ${{\rm{T}}_{Y \to {Y^*}}}\left( {i,j} \right){\rm{ = }}a$ if $i \ne j$ and ${{\rm{T}}_{Y \to {Y^*}}}\left( {i,j} \right){\rm{ = }}b$ if $i = j$, where $a,b > 0$ and $\left( C-1\right) a + b = 1$. For real-life data, although the type of label noise is complicated, its ${{\rm{T}}_{Y \to {Y^*}}}$ can be estimated through empirical distribution. Then, we have

\label{ass:AS}
\begin{assumption}
Assume the Markov chain: $X \to Y \to {Y^*}$ is hold. i.e., $X$ is independent of ${Y^*}$ conditioning on $Y$. Assume the transition distribution matrix  ${{\rm{T}}_{Y \to {Y^*}}}$ is invertible. i.e., ${\rm{det}}\left( {{{\rm{T}}_{Y \to {Y^*}}}} \right) \ne 0$.
\end{assumption}

From the Assumption \ref{ass:AS}, we can know that $  X  \Vbar {Y^ * }\mid Y $. The invertible transition distribution matrix is to emphasize that the any class in $\left\{ {1,2,...,C} \right\}$ can be polluted by noise, and each class can be transferred to every $C$ class with a certain possibility. This assumption is also to simulate the fact that every real label in the actual situation can be artificially incorrectly labeled as other classes.

\subsection{Representation leaning-based domain adaptation}
This paper focus on learning a domain-invariant representation for domain adaptation. The objective of representation learning-based domain adaptation methods is composed of three components \cite{gani06, wu10}. First, all training samples is mapped into the latent space $\mathcal{Z}$ to obtain the feature representation by a projection function (or feature extractor) $f$. The first is to minimize a metric which measures the difference between two probability distributions to align the distributions of two domains. The second part is to minimizing the source domain classification risk. The third is to minimize a regularization term. In short, the objective is formulated as
\begin{equation}
\begin{aligned}
\label{we123}
\mathop {\min }\limits_{f,h} {R_{{P_s}}}\left( {f,h} \right) + \alpha D\left( {P_s^{f\left( X \right)},P_t^{f\left( X \right)}} \right) + \Delta \left( {f,h} \right),
\end{aligned}
\end{equation}
where $D\left( \cdot \right) $ is the metric that measures the difference between two domain distributions, $\Delta $ is a regularization term to punish the parameters of the feature extractor $f$ and the classifier $h$, $\alpha$ is the corresponding hyperparameters, ${R_{{P_s}}}$ is the risk of the classifier $h$ over source domain samples based on the learned latent space $Z$:
\begin{equation}\label{12}
\begin{aligned}
{R_{{P_s}}}\left( f,h \right) = \int {{P_s}\left( X \right)} \left| {\eta \left( X \right) - h \left( {f \left( X \right)} \right)} \right|dX.
\end{aligned}
\end{equation}

In our setting, we only have access to the source domain samples with noisy labels. The aim is to learn an expected classifier ${h^*}$ that is robust to label noises. Thus, in our setting, the equation \ref{12} is written as
\begin{equation}\label{1w2}
\begin{aligned}
{R_{{P_s}}}\left( f,h \right) = \int {{P_s}\left( X \right)} \left| {Y^{*} - h \left( {f \left( X \right)} \right)} \right|dX.
\end{aligned}
\end{equation}

\begin{figure*}[tbh]	
	\begin{center}
		\centerline{\includegraphics[width=.9\textwidth]{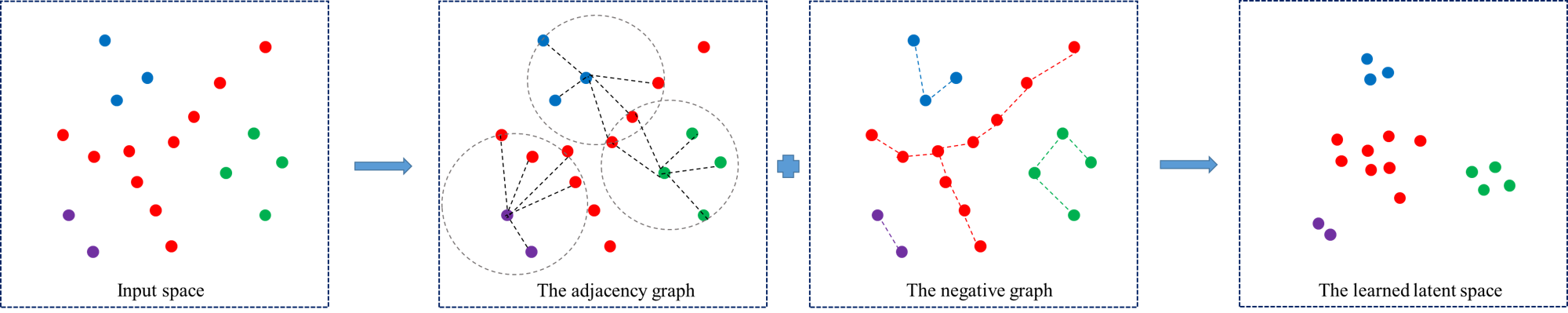}}
		\caption{A motivating example for the local preserving module. It contains an example for the adjacency graph and an example for the negative graph. We set $k=3$ in construct the adjacency graph. The number of the finally obtained clusters in the negative graph is 4. }
		\label{ag}		
	\end{center}	
	\vskip -0.35in
\end{figure*}

\section{Proposed method}
In this section, we introduce the proposed robust local preserving and global aligning network (RLPGA). RLPGA consists of four parts. The first part is the feature extractor $f$ that mapping the input dataset into the latent space to obtain the feature representation. The second part is the classifier to be learned based on the noisy labels of the source domain dataset. The third part is the local preserving module, which is to improve the robustness to the label noise based on preserving the local topological structure of the data in the input space. The fourth part is the global aligning module, which is to align the distributions of the source domain and the target domain. The overall framework of RLPGA is shown in Fig. \ref{fig:fram}.

\subsection{The informative-theoretic-based loss for classifier}
\label{ssub:1}
To learn a classifier $h$ with the noisy label for UDA, we employ the determinant based mutual information (DMI) \cite{kong2020} to measure the performance of the classifier. We maximize the DMI between the the output of $h$ and the noisy label ${Y^*}$. From the \cite{kong2020, xu2019l}, we can obtain the follows: 
\begin{definition} \label{def:DMI}(Determinant based Mutual Information)
	 For two discrete random variables ${X_1},{X_2}$, the determinant based mutual information between them is defined as
	\begin{equation}
	\begin{aligned}
	{\rm{DMI}}\left( {{X_1},{X_2}} \right) = \left| {\det \left( {{\rm{T}}_{{X_1},{X_2}}} \right)} \right|,
	\end{aligned}
	\end{equation}
	where ${\rm{T}}_{{X_1},{X_2}}$ is the matrix format of the joint distribution $\Pr \left( X_1, X_2 \right)$ over ${X_1}$ and ${X_2}$. 
\end{definition}

Therefore, let $Y^*$ be the one-hot vector, the measurement for the performance of the classifier $h$ is presented as ${\rm{DMI}}\left( {h\left(  \cdot  \right),{Y^*}} \right)$. As we can see, calculating ${\rm{DMI}}\left( {h\left(  \cdot  \right),{Y^*}} \right)$ need to obtain the joint distribution $\Pr \left( h\left(  \cdot  \right), Y^* \right)$. Because $h$ is also a random variable, therefore, we can obtain:
\begin{equation}\label{1}
\begin{aligned}
\Pr \left( {h\left(  \cdot  \right),{Y^ * }} \right) &= \int_X {\Pr \left( {h\left(  \cdot  \right),{Y^ * },X} \right)} d\left( X \right)\\
&= \int_X {\Pr \left( {h\left(  \cdot  \right),{Y^ * }\left| X \right.} \right)} P\left( X \right)d\left( X \right) \\
&= \int_X {\Pr \left( {h\left( X \right),{Y^ * }\left( X \right)} \right)} P\left( X \right)d\left( X \right)
\end{aligned}
\end{equation}
Also, we can obtain the Markov chain: $h\left( X \right) \leftarrow X \to {Y^ * }$. So, we can see:  $ {h\left( X \right)} \Vbar {Y^ * }\mid X $. Thus, during training process, a mini-batch of samples of source domain with their noisy labels are sampled and denoted as $\left\{ {\left( {{X_i},Y_i^*} \right)} \right\}_{i = 1}^N$. We denote the outputs of the classifier $h$ for these samples as the $N \times C$ matrix $ O $. Each row of $ O $ represents the output of a sample, and each column of $ O $ is a probability value over $C$ categories. We denote the ${Y^*}$ by a $0 - 1$ matrix $L$. Each row of $L$ is a one-hot vector and represents the label of the corresponding sample. We have ${{\rm{T}}_{h\left( X \right), Y^*}} = {{OL} \mathord{\left/
		{\vphantom {{OL} N}} \right.
		\kern-\nulldelimiterspace} N}$, Thus, ${\rm{DMI}}\left( {h\left(  X  \right),{Y^ * }} \right) = \left| {\det \left( {{OL} \mathord{\left/
		{\vphantom {{OL} N}} \right.
		\kern-\nulldelimiterspace} N} \right)} \right|$. 

To this end, the final loss function is defined as 
\begin{equation}
{L_{clf}} =  - \log \left( {\left| {\det \left( {\frac{1}{N}OL} \right)} \right|} \right)  + {\gamma  L_r}
\end{equation}
where $\gamma  > 0$ is the hyperparameter to weight the importance of the regularization item, the regularization item ${ L_r}$ is defined as
\begin{equation}
\label{221}
{L_r} = \frac{1}{N}\sum\limits_{i = 1}^N {H\left( {{O_{i,:}}} \right)}  - H\left( {\frac{1}{N}\sum\limits_{j = 1}^N {{O_{:,j}}} } \right)
\end{equation}
where the first term of equation \ref{221} aims at forcing the output of $h$ to be sparse enough to approximate an one hot vector, the second term of equation \ref{221} aims at forcing the output of $h$ to be equally distributed. The regularization item can also be interpreted as to constrain the source domain samples to be clustered into multiple clusters in a unsupervised manner.

\subsection{The local preserving module}
\label{ssub:3}
The local preserving module is to improve the robustness of RLPGA to label noise from the perspective of the learned feature representation. As we can see, the label noise of the source domain samples can cause a sample belonging to class $i$ to be annotated into class $j$. From the Fig. \ref {fig:fram}, we can see that these noisy labels will influence the measurement of the classifier, e.g., the loss ${L_{clf}}$. During the gradient propagation stage, the feature representation will be learned by mistake. From a geometrically intuitive point of view, this mistake can be regarded as pushing a $i$-th class sample from an area surrounded by $i$-th class samples to an area surrounded by $j$-th class samples. Therefore, one solution to improve this problem is to design a method that can maintain the local structure of the original input space in the learned latent feature space. To this end, inspired by \cite{li2020unsupervised}, we propose to excavate two kinds of relations. One is a similar relationship that should be preserved, the other is a dissimilar relationship that should be punished. We construct two weight matrices including an adjacency weight matrix and a negative weight graph to describe the above two relations.


Let ${X^s} = \left\{ {X_1^s,...,X_{{m_s}}^s} \right\}$ be the source domain dataset, where $X_i^s$ is a source domain sample and $m_s$ is the total number of samples in the source domain dataset. Let ${X^t} = \left\{ {X_1^t,...,X_{{m_t}}^t} \right\}$ be the target domain dataset, where $X_i^t$ is a target domain sample and $m_t$ is the total number of samples in the target domain dataset. Inspired by Locally Preserved Projection (LPP) \cite{heee2y}, $k$-nearest-neighbor method is introduced to construct the adjacency graph. For the input space of the source domain, if $X_j^s$ is one of the $k$ nearest neighbors of $X_i^s$ (or $X_i^s$ is one of the $k$ nearest neighbors of $X_j^s$), we build an edge between $X_i^s$ and $X_j^s$. We traverse all the input samples, and then get the adjacency graph, which consists of the vertexes (the source domain samples) and the edges. An example of the adjacency graph is shown in Fig. \ref{ag} and we set $k=5$. It is known that an instance and its $k$ nearest neighbors are very likely to belong to the same category. Thus, the adjacency graph can represent the local similar relationship of the input dataset. We also define $H_{pos}^s \in {R^{{m_s} \times {m_s}}}$ as adjacency weight matrix. $H_{pos}^s\left( {i,j} \right)$ can be regarded as the similarity between $X_i^s$ and $X_j^s$, which can be defined as: If $X_i^s \in {N_k}\left( {X_j^s} \right)$ or ${\rm{ }}X_j^s \in {N_k}\left( {X_i^s} \right)$, 
\begin{equation}
\label{222}
\begin{aligned}
H_{pos}^s\left( {i,j} \right) = \exp \left( {\frac{{ - {{\left( {d\left( {X_i^s,X_j^s} \right)} \right)}^2}}}{{{t_1}}}} \right)
\end{aligned}
\end{equation}
otherwise, $H_{pos}^s\left( {i,j} \right) = 0$. ${N_k}\left( {X_i^s} \right)$ is the $k$ nearest neighbors of $X_i^s$, $d\left(  \cdot  \right)$ is a distance metric, and parameter ${t_1}$ is a prespecified hyperparameter. As we can see, the adjacency weight matrix $H_{pos}^s$ can not only represent the connection state of any two input samples, but also the importance of any connected edge (or the similarity of any two input samples). Similarly, for target domain dataset, we can also obtain the corresponding adjacency weight matrix $H_{pos}^t\in {R^{{m_t} \times {m_t}}}$.


The negative weight matrix is to express the structure of dissimilar among input samples. We need to divide the samples of each domain into several clusters, and we suppose the samples that belong to different clusters are dissimilar. For source domain, if $X_j^s$ is the nearest neighbors of $X_i^s$, we build an edge between $X_i^s$ and $X_j^s$. We traverse all the input samples and finally construct a graph called negative graph, which consists of the vertexes (the source domain samples) and the edges. We definite that if there is a path between $X_i^s$ and $X_j^s$, then $X_i^s$ and $X_j^s$ belonging to the same cluster. We traverse the obtained graph and finally construct several clusters. An example of obtained clusters is shown in Fig. \ref{ag} and we obtain 3 clusters. Let $\left\{ {1,2, \ldots ,M} \right\}$ be cluster number and $M$ is the total number of clusters. Let $B \in {\mathbb{R}^{{m_s} \times 1}}$, and ${B_i} =m$, which means that sample $X_i^s$ belongs to the cluster $m$, $m \in \left\{ {1,2, \ldots ,M} \right\}$. We also define the negative weight matrix as $H_{neg}^s \in {R^{{m_s} \times {m_s}}}$. $H_{neg}^s\left( {i,j} \right)$ represents dissimilarity between $X_i^s$ and $X_j^s$, which can be defined as: If ${B_i} \ne {B_j}$, we have
\begin{equation}
\label{Eq:aa}
\begin{aligned}
H_{neg}^s\left( {i,j} \right) = \exp \left( {\frac{{{{-\left( {d\left( {X_i^s,X_j^s} \right)} \right)}^2}}}{{{t_1}}}} \right)
\end{aligned}
\end{equation}
otherwise, $H_{neg}^s\left( {i,j} \right) = 0$. The parameter ${t_1}$ is a prespecified hyperparameter. We can observe from the equation \ref{Eq:aa} that if $X_i^s$ and $X_j^s$ belong to different clusters, the closer the distance between them, the larger the value of $H_{neg}^s\left( {i,j} \right)$. This indicates that the boundary points of different clusters are important. Also, the negative weight matrix can not only indicate whether any two input samples belong to the same cluster, but also the dissimilarity of any two input samples. Similarly, as for target domain samples, we can also obtain a negative weight matrix $H_{neg}^t \in {R^{{m_t} \times {m_t}}}$.

To this end, let $H_{pn}^u = H_{pos}^u - H_{neg}^u$ and $u \in \left\{ {s,t} \right\}$, the objective of the local preserving module is denoted as:
\begin{equation}
\label{Eq:ab}
\begin{aligned}
\begin{array}{l}
\mathop {\min }\limits_f {Dis_{pn}}\left( f \right) = \\
\log \left( {1 + \sum\limits_{u \in \left\{ {s,t} \right\}} {\sum\limits_{i,j = 1}^{{m_u}} {\exp \left( {{{\left\| {f\left( {X_i^u} \right) - f\left( {X_j^u} \right)} \right\|}_2^2}H_{pn}^u\left( i,j\right) } \right)} } } \right)
\end{array}
\end{aligned}
\end{equation}
Also, the choice of distance $d\left(  \cdot  \right)$ in equation \ref{222} and \ref{Eq:aa} is based on different dataset. For higher dimensional data, we choise cosine distance, otherwise, choise Euclidean distance. 

We can see that minimizing equation \ref{Eq:ab} can make points that are near in the original space project closer to the latent space, and points that belong to different clusters in the original space project farther into the latent space. This can be regarded as preserving the local structure including the similarity structure and dissimilar structure. Based on the prior knowledge that samples belonging to the same category are likely gathered together and samples belonging to different categories are likely separated from each other. So, minimizing equation \ref{Eq:ab} can enhance the feature discriminability. As the training process continues, noisy label could make samples from some classes are mixed with samples from other classes. This can reduce the feature discriminability in the latent space. Thus, when aligning the distributions of two domains, some bad results will occur. However, the proposed local preserving method is to happen to force the features learned to be more discriminative, thereby improving the robustness of RLPGA to label noises.

\subsection{The global aligning module}
\label{ssub:2}
The global aligning module is to align the distributions between the source domain and target domain. The 'global' means the alignment operation is related to the whole distribution. Motivated by \cite{shen09}, we minimize the Wasserstein distance to align the two distributions. Specifically, for $\forall {P_r},{P_g} \in {\rm{Prob}}\left( X \right)$ and the corresponds support set ${\Sigma _r},{\Sigma _g}$, the $p{\rm{th}}$ Wasserstein distance can be defined as
\begin{equation}
\label{Eq:ad}
\begin{aligned}
{W_p}\left( {{P_r},{P_g}} \right) = {\left( {\mathop {\inf }\limits_{\varsigma \left( {X_a,X_b} \right) \in \Pi \left( {X_a,X_b} \right)} \int {c{{\left( {X_a,X_b} \right)}^p}d\varsigma } } \right)^{\frac{1}{p}}},
\end{aligned}
\end{equation}
where $X_a \in {\Sigma _r},X_b \in {\Sigma _g}$, $c\left( {X_a,X_b} \right)$ represents the distance of two patterns in ${\Sigma _r},{\Sigma _g}$ and $\Pi \left( {X_a,X_b} \right)$ denotes the set of all joint distributions $\varsigma \left( {X_a,X_b} \right)$ that satisfies ${P_r} = \int_y {\varsigma \left( {X_a,X_b} \right)d{X_b}} ,{P_g} = \int_x {\varsigma \left( {X_a,X_b} \right)d{X_a}} $. Based on Kantorovich-Rubinstein theorem \cite{san011}, the dual form of Wasserstein distance is written as
\begin{equation}
\label{Eq:af}
\begin{aligned}
{W_p}\left( {{P_r},{P_g}} \right) = \mathop {\sup }\limits_{{{\left\| \vartheta  \right\|}_L} \le 1} \mathop E\limits_{X_a \sim {P_r}} \left[ {\vartheta \left( X_a \right)} \right] - \mathop E\limits_{X_b \sim {P_g}} \left[ {\vartheta \left( X_b \right)} \right],
\end{aligned}
\end{equation}
where $\vartheta :X \to R$ is the 1-Lipschitz function and satisfies ${\left\| \vartheta  \right\|_L} = \mathop {\sup }\limits_{X_1 \ne X_2} {{\left| {\vartheta \left( X_1 \right) - \vartheta \left( X_2 \right)} \right|} \mathord{\left/
		{\vphantom {{\left| {\vartheta \left( X_1 \right) - \vartheta \left(  X_2 \right)} \right|} {\left| {X_1 -  X_2} \right|}}} \right.
		\kern-\nulldelimiterspace} {\left| {X_1-  X_2} \right|}} \le 1$. Also, the $\vartheta$ is called as the Critic and is implemented by an MLP.

To this end, the objective of the global aligning module is defined as,
\begin{equation}
\label{Eq:ah}
\begin{aligned}
\mathop {\min }\limits_f {W_p}\left( {P_s^{f\left( X \right)},P_t^{f\left( X \right)}} \right).
\end{aligned}
\end{equation}

\subsection{The final objective function}
Based on Subsection \ref{ssub:1}, \ref{ssub:3}, and \ref{ssub:2}, the overall objective function of RLPGA is formulated as following, and the training process of RLPGA is shown in Algorithm \ref{alg:RLPGA}. 
\begin{equation}
\label{Eq:hh}
\begin{aligned}
\begin{array}{l}
\mathop {\min }\limits_{f,h} {L_{clf}}\left( {f,h} \right) + \alpha Di{s_{pn}}\left( f \right) + \\
\;\;\;\;\;\;\;\;\;\;\;\;\;\;\;\;\;\;\;\;\;\beta {W_p}\left( {P_s^{f\left( X \right)},P_t^{f\left( X \right)}} \right)
+ \Delta \left( {f,h} \right)
\end{array}
\end{aligned}
\end{equation}

\begin{figure}[!t]
	\vskip -0.08in
	\label{alg:RLPGA}
	\renewcommand{\algorithmicrequire}{\textbf{Input:}}
	\renewcommand{\algorithmicensure}{\textbf{Output:}}
	\removelatexerror	
    \begin{algorithm}[H]
    	\caption{: RLPGA}
    	\begin{algorithmic}
    		\STATE {\bfseries Input:} source data $X^s$, target data ${X^t}$, minibatch size $m$, the Critic learning rate ${\alpha _1}$ in Wasserstein distance, the feature extractor learning rate ${\alpha _2}$, the classifier learning rate ${\alpha _3}$, hyperparameters $\alpha$, $\beta$, $\gamma $, and the number of neighbor points $k$.
    		\STATE Initialize the neural network parameters of the feature extractor $f$, the classifier $h$, and the Critic $\vartheta$ with random weights ${\omega _f}$, ${\omega _h}$, ${\omega _{\vartheta}}$.
    		\REPEAT
    		\STATE Sample minibatch samples from ${X^s}$ and ${X^t}$. Construct four weight matrices $H_{pos}^s$, $H_{neg}^s$, $H_{pos}^t$, $H_{neg}^t$.		
    		\FOR{$t=1$ {\bfseries to} $m$}
    		\STATE${Z^s} \leftarrow {f_{{\omega _f}}}\left( {{X^s}} \right),{Z^t} \leftarrow {f_{{\omega _f}}}\left( {{X^t}} \right)$ \\
    		${\omega _{wd}} \leftarrow {\omega _{wd}} + {\alpha _1}{\partial _{{\omega _{\vartheta}}}}{W_p}$ \\
    		\ENDFOR
    		\STATE ${\omega _h} \leftarrow {\omega _h} - {\alpha _3}{\partial _{{\omega _h}}}{L_{clf}}$ \\
    		${\omega _f} \leftarrow {\omega _f} - {\alpha _2}{\partial _{{\omega _f}}}\left[ {{L_{clf}} + \alpha Di{s_{pn}} + \beta {W_p} + \Delta} \right]$ 
    		\UNTIL ${\omega _f}$, ${\omega _h}$, ${\omega _{\vartheta}}$ converge.
    	\end{algorithmic}
    \end{algorithm}
\vskip -0.3in
\end{figure}
	
\subsection{Theoretical analysis}
We provide some theoretical analysis about the robustness and target risk on our proposed RLPGA. 

\begin{lemma} \label{lem:d} (Properties of DMI \cite{kong2020}).
	DMI is non-negative, symmetric, and information-monotone. Moreover, it is relatively invariant: for random variables ${X_1},{X_2}$ and ${X_3}$, if ${X_3}$ is independent of ${X_2}$ conditioning ${X_1}$, let ${{\rm{T}}_{{X_1} \to {X_3}}}$ be the matrix format of the joint distribution $\Pr \left( {{X_3}\left| {{X_1}} \right.} \right)$, then, the following holds
	\begin{equation}
	{\rm{DMI}}\left( {{X_2},{X_3}} \right) = {\rm{DMI}}\left( {{X_2},{X_1}} \right)\left| {\det \left( {{\rm{T}}_{{X_1} \to {X_3}}} \right)} \right|
	\end{equation}
\end{lemma}

\begin{theorem}
	\label{thm:1}
	For UDA with the noisy label, the proposed RLPGA is robust to label noise and the informative-theoretic-based loss is conducive to shrinking the upper bound of target risk ${R_{{P_t}}}$.
\end{theorem}
\begin{IEEEproof}
	For objective \ref{we123}, the first term, used to measure the performance of the classifier, is influenced by the noisy label, directly. From Lemma \ref{lem:d} , we can obtain
	\begin{equation}
	\label{Eq:the}
	\begin{aligned}
	\begin{array}{l}	
	{\rm{DMI}}\left( {h\left( \cdot \right),{Y^{*}}} \right) = {\rm{DMI}}\left( {h\left( \cdot \right),{Y}} \right)\left| {\det \left({{\rm{T}}_{{Y} \to {Y^*}}} \right)} \right|
	\end{array}
	\end{aligned}
	\end{equation}
	where $Y^*$ is the noisy version of the ground truth label $Y$. For every two classifiers $h_1$ and $h_2$, we can see that the necessary and sufficient conditions for ${\rm{DMI}}\left( {{h_1}\left( \cdot \right),Y} \right) > {\rm{DMI}}\left( {{h_2}\left( \cdot \right),Y} \right)$ is ${\rm{DMI}}\left( {{h_1}\left( \cdot \right),Y^*} \right) > {\rm{DMI}}\left( {{h_2}\left( \cdot \right),Y^*} \right)$. In this paper, we propose to use the informative-theoretic-based loss to measure the performance of the classifier, we thus obtain that the necessary and sufficient conditions for $L_{clf}\left( {f, {h_1}\left( \cdot \right), Y} \right) > L_{clf}\left( {f, {h_2}\left( \cdot \right),Y} \right)$ is $L_{clf}\left( {f, {h_1}\left( \cdot \right),Y^*} \right) > L_{clf}\left( {f, {h_2}\left( \cdot \right),Y^*} \right)$. Therefore, we can obtain that the measurement based on noisy labels is consistent with the measurement based on clean labels. To this end, we conclude that RLPGA is robust to label noise.
	
    Let $h_1$ be the learned classifier, from \cite{shen09}, we can know that the target error of the UDA can be bounded by the follows:
    \begin{equation}
    \label{Eq:tthe}
    \begin{aligned}
    {R_{{P_t}}}\left( {{h_i}} \right) \le {R_{{P_s}}}\left( {{h_i}} \right) + 2{W_p}\left( {P_s^{f\left( X \right)},P_t^{f\left( X \right)}} \right) + \psi 
    \end{aligned}
    \end{equation}
    As we can see, if the objective \ref{we123} is minimized, the first term ${R_{{P_s}}}\left( {{h_i}} \right)$ is equal to $0$. When the label is the noisy version, for traditional loss function such as cross entropy loss, even if ${R_{{P_s}}}\left( {{h_i}} \right)$ is equal to $0$, the value of ${R_{{P_s}}}\left( {{h_i}} \right)$ under the clean ground truth label is greater than 0. But for our proposed loss function, the result under the noisy labels is consistent with result under the clean labels. Thus, we can obtain that the upper bound of ${R_{{P_t}}}\left( {f,{h_1}} \right)$ is less than the upper bound of ${R_{{P_t}}}\left( {f,{h_2}} \right)$.

\end{IEEEproof}

\begin{proposition}
	\label{thm:2}
	When $Z$ is a deterministic function of $X$, minimizing the equation \ref{Eq:ab} is also conducive to minimizing the target risk ${R_{{P_t}}}\left( {f,h} \right)$.
\end{proposition}
\begin{IEEEproof}
	When $Z$ is a deterministic function of $X$, we can obtain $\Pr \left( {Z\left| {{X^u}} \right.} \right), u \in \left\{ {s,t} \right\}$ is Dirac. Therefore, ${R_{{P_t}}}$ can also be rewritten as
	\begin{equation}
	\label{Eq:hddedk}
	\begin{array}{l}
	{R_{{P_t}}}\left( {f,h} \right)\\
	= \int {P_t^X\left( X \right)\left| {h\left( {f\left( X \right)} \right) - {Y^t}} \right|d\left( X \right)} \\
	\buildrel\textstyle.\over= \int {P_t^{f\left( X \right)}\left( Z \right)\left| {h\left( Z \right) - {Y^t}} \right|d\left( Z \right)} \\
	= {R_{{P_s}}}\left( {f,h} \right) - \int {P_s^{f\left( X \right)}\left( Z \right)\left| {h\left( Z \right) - {Y^s}} \right|d\left( Z \right)} \\
	\;\;\;\;\;\;\;\;\;\;\;\;\;\;\;\;\;\;\;\;\;\;\;\;\;\;\;+ \int {P_t^{f\left( X \right)}\left( Z \right)\left| {h\left( Z \right) - {Y^t}} \right|d\left( Z \right)} \\
	= {R_{{P_s}}}\left( {f,h} \right) + \int {P_t^{f\left( X \right)}\left( Z \right)\left( {{\Psi _t} - {\Psi _s}} \right)d\left( Z \right)} \\
	\;\;\;\;\;\;\;\;\;\;\;\;\;\;\;\;\;\;\;\;\;+ \int {\left( {P_t^{f\left( X \right)}\left( Z \right) - P_s^{f\left( X \right)}\left( Z \right)} \right){\Psi _s}d\left( Z \right)} 
	\end{array}
	\end{equation}
	where ${\Psi _u}\left( Z \right) = \left| {h\left( Z \right) - Y^u} \right|$, $u = \left\{ {s,t} \right\}$. Compared equation \ref{Eq:hddedk} with equation \ref{Eq:hh}, we can know that minimizing the first term and the third term in equation \ref{Eq:hh} corresponds to the minimizing first and third terms in equation \ref{Eq:hddedk}. It is impossible to directly minimize the second term of equation \ref{Eq:hddedk} in equation \ref{Eq:hh}. However, this paper mainly focuses on the covariance shift, so, we have $Y^s=Y^t$. The first term of equation \ref{Eq:hh} can increase the ability of the classifier $h$ to correctly classify the source domain samples. Minimizing equation \ref{Eq:ab} can increase the feature discriminability of both source domain samples and target domain samples, and the label information is related to the feature discriminability. So, minimizing equation \ref{Eq:ab} is conducive to make the learned classifier $h$ to predict the true label of the target domain samples. Therefore, we can conclude that minimizing equation \ref{Eq:ab} is conducive to minimizing the second term equation \ref{Eq:hddedk} thus is also conducive to minimizing target risk ${R_{{P_t}}}\left( {f,h} \right)$.

\end{IEEEproof}

\section{Experiments}
We conduct experiments on one synthetic data and four benchmark datasets. Also, due to space constraints, additional experiments of three datasets can be found in Appendix. Besides, we put the deepgoing analysis of the discrepancy metric and the training time analysis in Appendix. The main lines behind the experiment section are as follows. In Subsection \ref{5.3}, we take experiments on the synthetic dataset is to show that the gradient of our proposed model is stable and can converge during the training process under different scales of label noise. In Subsection \ref{sec:ccomp}, we first conducted conventional unsupervised domain adaptation experiments of transfer tasks on benchmark datasets, and the reported tables show the experimental results when the source domain samples have ground truth labels. This subsection is to verify that our method also has performance advantages in the absence of noise. In Subsection \ref{dcomp}, we show the experimental results when the labels of source domain samples are polluted by different noise ratios. This is to verify that our method is not only suitable for learning with label noise, but also for learning without label noise. In Subsection \ref{as}, we show the experimental results of the ablation study. This is to verify that both the informative-theoretic-based loss and the local preserving module can improve the robustness to the label noise. In Subsection \ref{tf}, we perform several experiments to study the influence of the hyper-parameters in our proposed RLPGA. In Subsection \ref{fv}, we show the feature transferability and the feature discriminability of our proposed RLPGA, intuitively. Also, in Appendix, we have given the experimental results in the case of random noise under the Digits dataset.


\begin{figure}[!t]
	\centering		
	\includegraphics[width=2.5in]{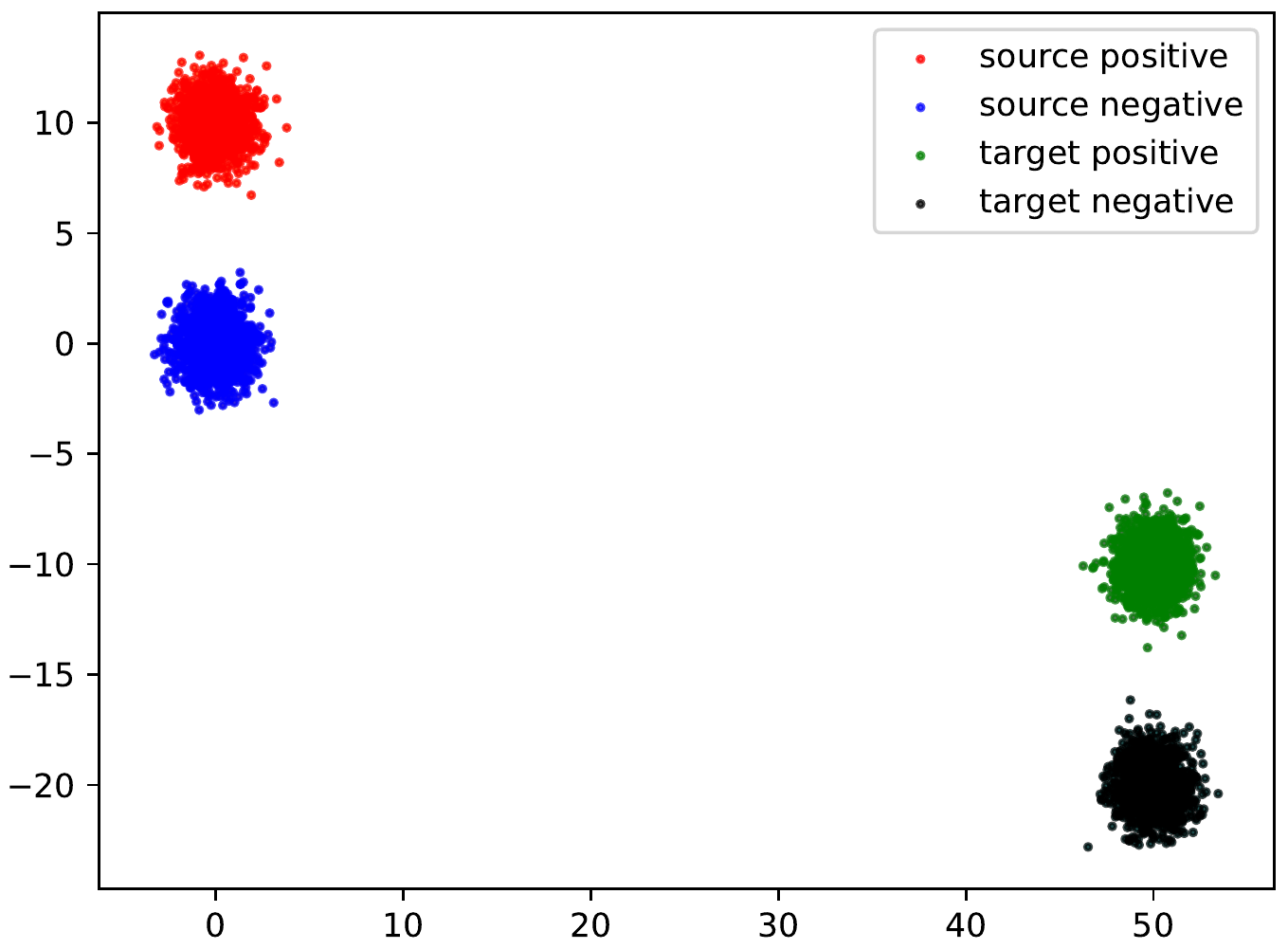}
	\captionsetup{justification=centering}
	\caption{Synthetic dataset}
	\label{fig:112233}			
	\vskip -0.1in
\end{figure}



\textbf{Synthetic dataset} is the combination of source and target domain samples. Each domain includes 1000 positive samples and 1000 negative samples. The synthetic dataset is visualized in Fig. \ref{fig:112233}. We can see that the positive samples and negative samples of the source domain are represented by red dots and blue dots, and the positive samples and negative samples of the target domain are represented by green dots and black dots. \textbf{Office-Caltech10 dataset} \cite{gong201} includes four domains, \textit{e.g.,} Amazon (A), Webcam (W), DSLR (D). and Caltech (C). We adopt 4096-dimensional DeCaf feature \cite{dona2014} for the samples in Office+Caltech10 dataset. \textbf{Office31 dataset} \cite{saenko2010} includes three domains, \textit{e.g.,} Amazon (A), Webcam (W) and DSLR (D). \textbf{Office-Home dataset} \cite{ven2017deep} includes four domains, \textit{e.g.,} Artistic images (Ar), Clip Art (Cl), Product images (Pr), and Real-World images (Rw). \textbf{Digits datasets} \cite{gani06} includes two dataset including MINIST dataset and USPS dataset.

\begin{table*}[!h]
	\caption{Performance (accuracy) on Office+Caltech10 dataset with DeCaf features}
	\label{tab:bb}
	\setlength{\tabcolsep}{1.9pt}
	\vskip 0.005in
	\begin{center}
		\begin{small}
			\begin{tabular}{c|ccccccccccccc}
				\hline
				Domains & A$\to$C & A$\to$D & A$\to$W & W$\to$A & W$\to$D & W$\to$C & D$\to$A & D$\to$W & D$\to$C & C$\to$A & C$\to$W & C$\to$D & Average \\
				\hline
				\hline
				TJM & 84.3 & 76.4 & 71.9 & 87.6 & \textbf{100} & 83.0 & 90.3 & 99.3 & 83.8 & 88.8 & 81.4 & 84.7 & 86.0 \\
				SCA & 78.8 & 85.4 & 75.9 & 86.1 & \textbf{100} & 74.8 & 90.0 & 98.6 & 78.1 & 89.5 & 85.4 & 87.9 & 85.9 \\
				ARTL & 87.4 & 85.4 & 88.5 & 92.3 & \textbf{100} & 88.2 & 92.7 & \textbf{100} & 87.3 & 92.4 & 87.8 & 86.6 & 90.7 \\
				JGSA & 84.9 & 88.5 & 81.0 & 90.7 & \textbf{100} & 85.0 & 92.0 & 99.7 & 86.2 & 91.4 & 86.8 & 93.6 & 90.0 \\
				CORAL & 83.2 & 84.1 & 74.6 & 81.2 & \textbf{100} & 75.5 & 85.5 & 99.3 & 76.8 & 92.0 & 80.0 & 84.7 & 84.7 \\
				DMM & 84.8 & 92.4 & 84.7 & 86.5 & 98.7 & 81.7 & 90.7 & 99.3 & 83.3 & 92.4 & 87.5 & 90.4 & 89.4\\
				\hline
				\hline
				AlexNet & 83.0 & 87.4 & 79.5 & 83.8 & \textbf{100} & 73.0 & 87.1 & 97.7 & 79.0 & 91.9 & 83.7 & 87.1 & 86.1 \\
				DDC & 85.0 & 89.0 & 86.1 & 84.9 & \textbf{100} & 78.0 & 89.5 & 98.2 & 81.1 & 91.9 & 85.4 & 88.8 & 88.2 \\
				DAN & 84.1 & 91.7 & 91.8 & 92.1 & \textbf{100} & 81.2 & 90.0 & 98.5 & 80.3 & 92.0 & 90.6 & 89.3 & 90.1 \\
				MMD & 88.6 & 90.5 & 91.6 & 92.2 & \textbf{100} & 88.6 & 90.1 & 98.9 & 87.8 & 93.1 & 91.6 & 91.2 & 92.0 \\
				DANN & 87.8 & 82.5 & 77.8 & 82.9 & \textbf{100} & 81.3 & 84.7 & 98.9 & 82.1 & 93.3 & 89.5 & 91.2 & 87.7 \\
				DCORAL & 86.2 & 91.2 & 90.5 & 88.4 & \textbf{100} & 88.6 & 85.8 & 97.9 & 85.4 & 93.0 & 92.6 & 89.5 & 90.8 \\
				MEDA & 87.4 & 88.1 & 88.1 & \textbf{99.4} & 99.4 & 93.2 & 93.2 & 97.6 & 87.5 & 93.4 & 95.6 & 91.1 & 92.8 \\
				WDGRL & 86.9 & 93.7 & 89.5 & 93.7 & \textbf{100} & 89.4 & 91.7 & 97.9 & 90.2 & 93.5 & 91.6 & 94.7 & 92.7 \\
				SWD & 85.1 & 92.3 & 89.5 & 92.2 & \textbf{100} & 88.1 & 90.9 & 97.4 & 91.6 & 92.9 & 90.7 & 92.9 & 92.0 \\
				\hline
				\hline
				\textbf{RLPGA} & \textbf{96.7} & \textbf{96.5} & \textbf{100} & 96.8 & \textbf{100} & \textbf{93.5} & \textbf{93.7} & 93.7 & \textbf{93.5} & \textbf{97.5} & \textbf{100} & \textbf{98.2} & \textbf{96.7} \\
				\hline
			\end{tabular}
		\end{small}
	\end{center}
	\vskip -0.2in
\end{table*}

\subsection{Compared Approaches}
In this paper, we mainly compare our proposed RLPGA with both traditional learning methods and deep domain adaptation methods. The specific transfer task classification accuracy and average classification accuracy are reported. All experimental results of the compared methods in our submission are quoted from their respective original papers. To ensure the fairness and authenticity of the experimental results, we have not reproduced the experimental results of the compared methods on the data set that did not appear in the original article.

\textbf{Traditional learning methods}:  Transfer Joint Matching (TJM) \cite{cao2018unsupervised}, Scatter Component Analysis (SCA) \cite{ghifary2016}, Adaptation Regularization (ARTL) \cite{long2013ada}, Joint Geometrical and Statistical Alignment (JGSA) \cite{zhang2017joint}, CORrelation Alignment (CORAL) \cite{sun201n}, Distribution Matching Machine (DMM) \cite{cao2018unsupervised}. 

\textbf{Deep domain adaptation methods}: AlexNet \cite{krizhevsky2012}, ResNet-50 \cite{he2016deep}, Deep Domain Confusion (DDC) \cite{tzeng2014}, Deep Adaptation Network (DAN) \cite{long2015}, Maximum Mean Discrepancy Metric (MMD) \cite{gre01}, Domain Adversarial Neural Network (DANN) \cite{gani06}, Deep Correlation Alignment (DCORAL) \cite{sun2016}, Adversarial Discriminative Domain Adaptation (ADDA) \cite{tzeng2017al}, Joint Adaptation Network (JAN) \cite{long2017deep}, Multi-Adversarial Domain Adaptation (MADA) \cite{pei2018multi}, Similarity Network (SimNet) \cite{pinheiro2018ud}, Generate to Adapt (GTA) \cite{sankn2018generate}, Deep Adversarial Attention Alignment (DAAA) \cite{kang2018deep}, Conditional Domain Adversarial Network (CDAN) \cite{long2018conditional}, Mainfold Embedded Distribution Alignment (MEDA) \cite{wang2018ual}, Batch Spectral Penalization (BSP) \cite{chen2019}, Wasserstein Distance Guided Representation Learning (WDGRL) \cite{shen09}, Contrastive Adaptation Network (CAN) \cite{kang2019contrastive}, Certainty Attention based Domain Adaption (CADA) \cite{kurmi201}, Sliced Wasserstein Discrepancy (SWD) \cite{lee2019sliced}.

\subsection{Implementation} \label{sec:implement}
All compared models in our experiments are implemented with TensorFlow and optimized by Adam optimizer. For each approach, all hyper-parameters are fixed, the batch size is set to 64 with 32 samples from each domain, the learning rate is set to ${\rm{1}}{{\rm{0}}^{{\rm{ - 4}}}}$, the projection function $f$, the classifier $h$, and the Critic to approximate the Wasserstein distance are all set as MLP network, in which the activation function is set as Relu. Also, a softmax function is attached behind the classifier $h$ to obtain a probabilistic output.

For the Synthetic dataset, the projection function $f$ is approximated by a MLP network with one hidden layer of 20 nodes, the Critic in Wasserstein distance is approximated by a MLP network with two layers of 20 and 1 nodes. For Amazon and Email datasets, the projection function $f$ is approximated by a MLP network with one hidden layer of 500 nodes, the Critic is approximated by a MLP network with two layers of 100 and 1 nodes. For Office-Caltech10, Office31, Office-Home, and Digits datasets, the projection function $f$ is approximated by a MLP network with two hidden layers of 500 and 100 nodes, the Critic is approximated by a MLP network with two layers of 100 and 1 nodes. Note that the deep representations obtained by 50-layer ResNet are used as features for Office31, Office-Home, and Digits datasets.

To verify the robustness of our proposed RLPGA, we conduct experiments under different ratio noise. We list the explicit noise transition matrices as following, and for all experiments, $r$ ranges from $\left\{ {0.2,0.4,0.6} \right\}$. For Synthetic dataset, case (1):


\begin{equation}
\begin{aligned}
{P_{Y{Y^*}}} = \left[ {\begin{array}{*{20}{c}}
	1&0\\
	r&{1 - r}
	\end{array}} \right].
\end{aligned}
\end{equation}
For the rest of dataset, case (2):
\begin{equation}
\begin{aligned}
\left[ {\begin{array}{*{20}{c}}
	1&{}&{}&{}&{}&{}&{}&{}\\
	{}& \cdots &{}&{}&{}&{}&{}&{}\\
	r&{}&{1 - r}&{}&r&{}&{}&{}\\
	{}&{}&{}& \cdots &{}&{}&{}&{}\\
	{}&{}&{}&{}&{1 - r}&{}&r&{}\\
	{}&{}&{}&{}&{}&1&{}&{}\\
	{}&{}&{}&{}&{}&{}& \cdots &{}\\
	{}&r&{}&{}&{}&{}&{}&{1 - r}
	\end{array}} \right]
\end{aligned}
\end{equation}

There are four hyperparameters in our proposed RLPGA, that is $\alpha$, $\beta$, $\gamma $, and $k$. For the synthetic dataset, we set $\alpha=1$, $\beta=0.1$, $\gamma=1$, and $k=3$. For Office-Caltech10 dataset, we set $\alpha=1$, $\beta=10$, $\gamma=1$, and $k=3$. For Office31 dataset, we set $\alpha=1$, $\beta=0.1$, $\gamma=1$, and $k=3$. For Office-Home dataset, we set $\alpha=1$, $\beta=10^{3}$, $\gamma=1$, and $k=3$. For Digits dataset, we set $\alpha=1$, $\beta=10$, $\gamma=1$, and $k=3$. For Email Spam Filtering dataset, we set $\alpha=1$, $\beta=10^{-2}$, $\gamma=0.1$, and $k=3$. For Amazon Review dataset, we set $\alpha=1$, $\beta=1$, $\gamma=10$, and $k=3$.


\begin{figure*}[!t]
	\vskip 0.1in
	\centering		
	\includegraphics[scale=0.8]{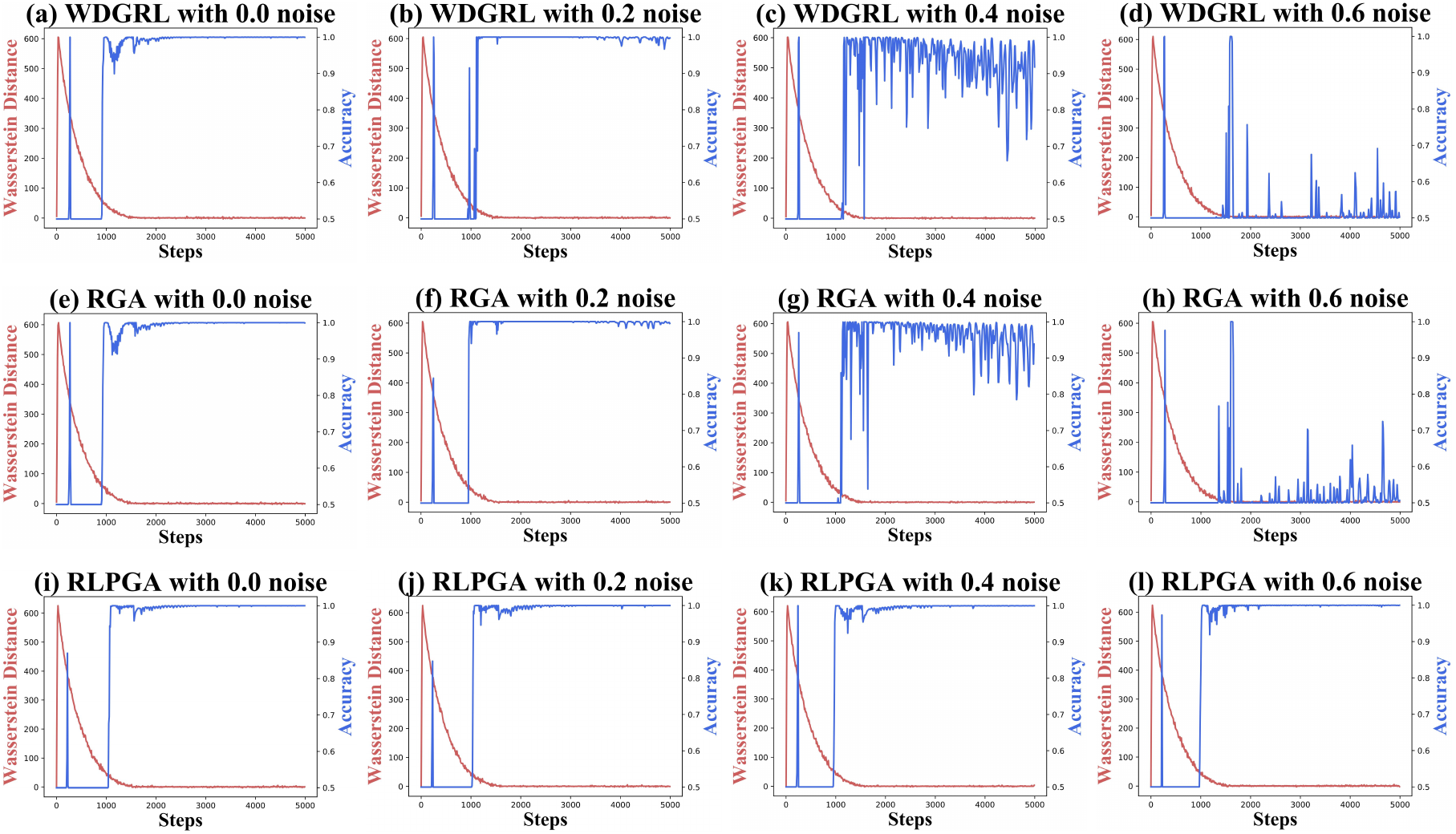}
	\captionsetup{justification=centering}
	\caption{The Wasserstein distance and test classification accuracy of WDGRL, RGA and RLPGA during each iteration}
	\label{fig:syplot}			
	\vskip -0.2in	
\end{figure*}	

\subsection{Experimental results on synthetic dataset} \label{5.3}
We mainly compare our proposed RLPGA with RGA and WDGRL. Specifically, in RGA, the hyperparameter $\alpha$ is set to 0, which aims to eliminate the impact of the proposed two weight graphs. Fig. \ref{fig:syplot} shows the experimental results of the synthetic dataset for three methods including WDGRL, RGA, and our proposed RLPGA. Each row in Fig. \ref{fig:syplot} represents a different label noise ratio, and each column represents a different method. We record the value of the Wasserstein distance and the test classification accuracy during each iteration of the training process. As we can see, the Wasserstein distances in all three methods converge when ${\rm{step}} > 2000$, and the convergence curve is very smooth. So, we can conclude that all three methods show the gradient priority. Compared RGA with WDGRL, when the noise ratio is 0, we can see that the accuracy curve of RGA and WDGRL is almost the same. However, as the proportion of noise increases, the accuracy curve of WDGRL is obviously more oscillating than the accuracy curve of RGA. As the noise ratio goes to 0.6, the accuracy curve of RGA and WDGRL is almost the same again. This indicates that the proposed robust informative theoretic-based loss function is effective to reduce the impact of label noise to a certain degree. Compared RLPGA with WDGRL and RGA, we can see that the accuracy curve of RLPGA is obviously more stable than the accuracy curves of WDGRL and RGA under all noise ratios. In particular, in the case of all 4 noise ratios, the oscillation of the RLPGA accuracy curves is very weak, and the accuracy reaches 1. This demonstrates that the proposed two weight graphs are effective to reduce the impact of label noise and can promote the stability of the training process.

\begin{table}[!h]
	\caption{Performance (accuracy) on Office31 dataset}
	\label{tab:dd}
	\setlength{\tabcolsep}{2pt}
	\begin{center}
		\begin{small}
			\begin{tabular}{c|ccccccc}
				\hline
				Domains & A$\to$D & A$\to$W & D$\to$A & D$\to$W & W$\to$A & W$\to$D & Average \\
				\hline
				\hline
				ResNet-50 & 68.9 & 68.4 & 62.5 & 96.7 & 60.7 & 99.3 & 76.1 \\
				DAN & 78.6 & 80.5 & 63.6 & 97.1 & 62.8 & 99.6 & 80.4 \\
				DANN & 79.7 & 82.0 & 68.2 & 96.9 & 67.4 & 99.1 & 82.2 \\
				ADDA & 77.8 & 86.2 & 69.5 & 96.2 & 68.9 & 98.4 & 82.8 \\
				JAN & 84.7 & 85.4 & 68.6 & 97.4 & 70.0 & 99.8 & 84.3 \\
				MADA & 87.8 & 90.0 & 70.3 & 97.4 & 66.4 & 99.6 & 85.3 \\
				SimNet & 85.2 & 88.6 & 73.4 & 98.2 & 71.6 & 99.7 & 86.1 \\
				GTA & 87.7 & 89.5 & 72.8 & 97.9 & 71.4 & 99.8 & 86.5 \\
				DAAA & 88.8 & 86.8 & 74.3 & \textbf{99.3} & \textbf{73.9} & \textbf{100.0} & 87.2 \\
				CDAN & 93.4 & 93.1 & 71.0 & 98.6 & 70.3 & \textbf{100.0} & 87.7 \\
				MEDA & 86.2 & 85.9 & 72.3 & 97.4 & 73.4 & 99.4 & 85.8 \\
				CAN & 81.5 & \textbf{99.7} & 85.5 & 65.9 & 63.4 & 98.2 & 82.4 \\
				CADA & 95.6 & 97.0 & 71.5 & \textbf{99.3} & 73.1 & \textbf{100.0} & 89.4 \\
				SWD & 83.5 & 82.5 & 85.7 & 88.9 & 72.5 & 96.4 & 84.9 \\
				\hline
				\hline
				\textbf{RLPGA} & \textbf{97.3} & 97.2 & 74.8 & 97.8 & 73.3 & \textbf{100.0} & \textbf{90.1} \\
				\hline
			\end{tabular}
		\end{small}
	\end{center}
	\vskip -0.2in
\end{table}

\begin{table*}[!h]
	\caption{Performance (accuracy) on Office-Home dataset}
	\label{tab:ee}
	\setlength{\tabcolsep}{1.9pt}
	\vskip 0.005in
	\begin{center}
		\begin{small}
			\begin{tabular}{c|ccccccccccccc}
				\hline
				Domains & Ar$\to$Cl & Ar$\to$Pr & Ar$\to$Rw & Cl$\to$Ar & Cl$\to$Pr & Cl$\to$Rw & Pr$\to$Ar & Pr$\to$Cl & Pr$\to$Rw & Rw$\to$Ar & Rw$\to$Cl & Rw$\to$Pr & Average \\
				\hline
				\hline
				ResNet-50 & 34.9 & 50.0 & 58.0 & 37.4 & 41.9 & 46.2 & 38.5 & 31.2 & 60.4 & 53.9 & 41.2 & 59.9 & 46.1 \\
				DAN & 43.6 & 57.0 & 67.9 & 45.8 & 56.5 & 60.4 & 44.0 & 43.6 & 67.7 & 63.1 & 51.5 & 74.3 & 56.3 \\
				DANN & 45.6 & 59.3 & 70.1 & 47.0 & 58.5 & 60.9 & 46.1 & 43.7 & 68.5 & 63.2 & 51.8 & 76.8 & 57.6 \\
				JAN & 45.9 & 61.2 & 68.9 & 50.4 & 59.7 & 61.0 & 45.8 & 43.4 & 70.3 & 63.9 & 52.4 & 76.8 & 58.3 \\
				CDAN & 49.0 & 69.3 & 74.5 & 54.4 & 66.0 & 68.4 & 55.6 & 48.3 & 75.9 & 68.4 & 55.4 & 80.5 & 63.8 \\
				CDAN+E & 50.7 & 70.6 & 76.0 & 57.6 & 70.0 & 70.0 & 57.4 & 50.9 & 77.3 & 70.9 & 56.7 & 81.6 & 65.8 \\
				BSP+DANN & 51.4 & 68.3 & 75.9 & 56.0 & 67.8 & 68.8 & 57.0 & 49.6 & 75.8 & 70.4 & 57.1 & 80.6 & 64.9 \\
				BSP+CDAN & 52.0 & 68.6 & 76.1 & \textbf{58.0} & \textbf{70.3} & 70.2 & \textbf{58.6} & 50.2 & 77.6 & \textbf{72.2} & 59.3 & 81.9 & 66.3 \\
				\hline
				\hline
				\textbf{RLPGA} & \textbf{54.3} & \textbf{73.3} & \textbf{76.9} & 57.7 & 69.9 & \textbf{71.2} & 57.6 & \textbf{51.4} & \textbf{79.4} & 71.9 & \textbf{60.2} & \textbf{82.5} & \textbf{67.2} \\
				\hline
			\end{tabular}
		\end{small}
	\end{center}
	\vskip -0.2in
\end{table*}

\begin{table}[!h]
	\caption{Performance (accuracy) on Digits dataset}
	\label{tab:gg}
	\setlength{\tabcolsep}{1.9pt}
	\vskip 0.005in
	\begin{center}
		\begin{small}
			\begin{tabular}{c|ccc}
				\hline
				Domains & MNIST$\to$USPS & USPS$\to$MNIST & Average \\
				\hline
				\hline
				DANN & 90.4 & 94.7 & 92.6 \\
				ADDA & 89.4 & 90.1 & 89.8 \\
				UNIT & 96.0 & 93.6 & 94.8 \\
				CyCADA & 95.6 & 96.5 & 96.1 \\
				CDAN & 93.9 & 96.9 & 95.4 \\
				CDAN+E & 95.6 & 98.0 & 96.8 \\
				BSP+DANN & 94.5 & 97.7 & 96.1 \\
				BSP+ADDA & 93.3 & 94.5 & 93.9 \\
				BSP+CDAN & 95.0 & 98.1 & 96.6 \\
				SWD & \textbf{98.1} & 97.1 & 97.6 \\
				\hline
				\hline
				\textbf{RLPGA} & 97.2 & \textbf{98.6} & \textbf{97.9} \\
				\hline
			\end{tabular}
		\end{small}
	\end{center}
	\vskip -0.2in
\end{table}

\subsection{Conventional comparisons on transfer task} \label{sec:ccomp}
We conducted conventional unsupervised domain adaptation experiments of transfer tasks on benchmark datasets, and the reported tables show the experimental results when the source domain samples have ground truth labels. We compare the performances of different methods based on specific transfer task classification accuracy and average classification accuracy.

\textit{1) Comparisons on the Office+Caltech10 dataset with DeCaf features:} From Table \ref{tab:bb}, we observe that the average classification accuracy of RLPGA is 96.7\%, which are 3.9\% higher
than the best among the 21 benchmark domain adaptation
methods, especially, 4.0\% higher than WDGRL, and 5.9\%
higher than DCORAL. As for specific transfer task classification accuracy, RLPGA achieves the best results on 10
specific transfer tasks. Also, we can observe that the best results among the 12 specific transfer tasks also appear in
the deep domain adaptation methods compared with the
traditional learning methods, and the best results are more
likely to appear in the last line. The improvements of our Proposed RLPGA in this dataset are significant. 

\textit{2) Comparisons on the Office31 dataset:} From Table \ref{tab:dd}, we
observe that the average classification accuracy of RLPGA
is 90.1\%, which is 0.7\% higher than the best among the
compared 14 domain adaptation methods, especially, 5.2\%
higher than SWD, and 7.9\% higher than DCORAL. As for
specific transfer tasks, RLPGA achieves the best results on
2 specific transfer tasks and obtains comparable results on
1 specific transfer task. Also, we observe that the results of
domain adaptation methods are all better than the results of
the source-only based method, i.e., ResNet-50 \cite{he2016deep}. Overall, the improvements of our Proposed RLPGA in this dataset are not significant. 

\textit{3) Comparisons on the Office-Home dataset:} From Table \ref{tab:ee}, we can know that RLPGA achieves the best results on most
tasks. For example, the average classification accuracy of
RLPGA is 67.2\%, which is 0.9\% higher than the best among
the compared 8 domain adaptation methods, 9.6\% higher
than DANN, and 21.1\% higher than ResNet-50. As for specific transfer task classification accuracy, RLPGA achieves
the best results on 8 of 12 specific transfer tasks and obtains
comparable results on the other 4 specific transfer tasks. The improvements of RLPGA in this dataset are significant.

\textit{4) Comparisons on the Digits dataset:} From Table \ref{tab:gg}, we observe that the average accuracy of RLPGA outperforms
all other methods. The experimental results are consistent
with the comparisons of the previous experiments. The improvements of our Proposed RLPGA in this data are not significant.

In general, we can draw the following conclusions for
the conventional comparisons on transfer task: 1) Deep domain adaptation methods are more effective than traditional
learning methods; 2) The adversarial based methods are
more effective than metric-based methods; 3) The learned
latent feature representation of our proposed RLPGA is the
most discriminative. 

\begin{figure*}[!h]
	\vskip 0.2in
	\begin{center}
		\centering
		{\includegraphics[scale=0.8]{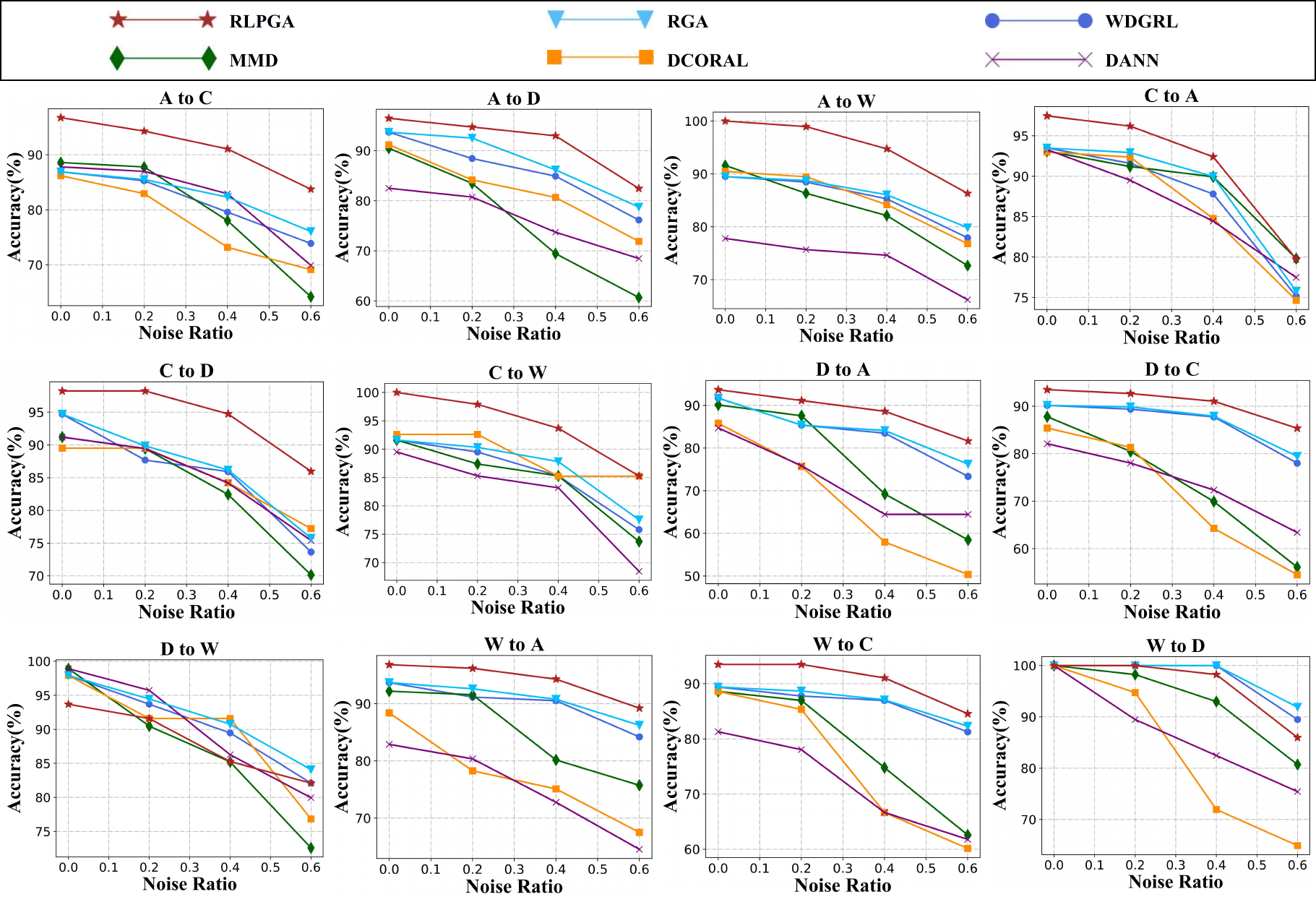}}
		\caption{Robustness evaluation on Office+Caltech10 dataset with DeCaf features}
		\label{fig:bb}
	\end{center}
	\vskip -0.2in
\end{figure*}

\begin{figure*}[!h]
	\vskip 0.2in
	\begin{center}
		\centering{\includegraphics[width=0.85\textwidth]{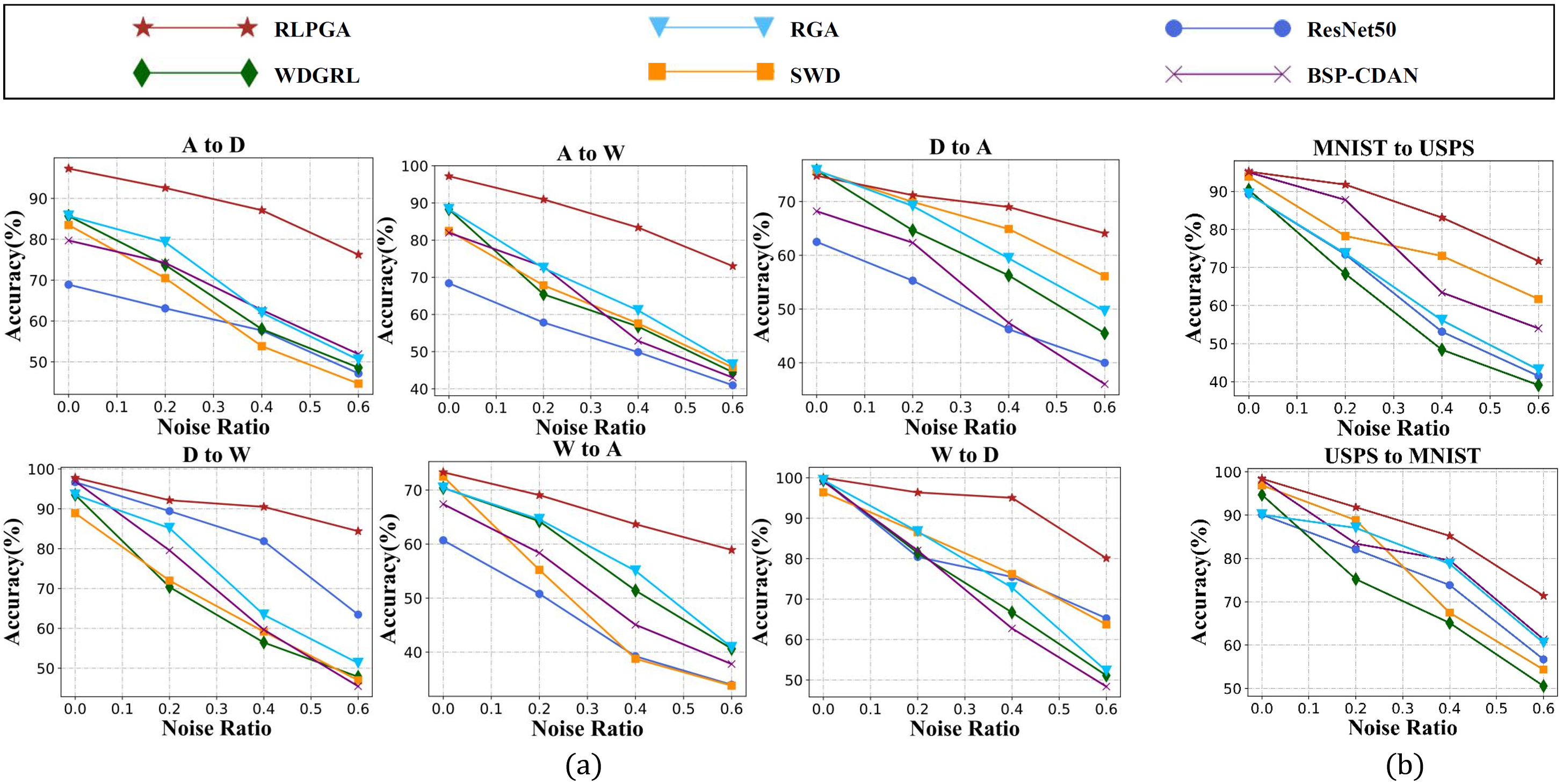}}
		\caption{Robustness evaluation on (a) Office31 dataset; (b) Digits dataset}
		\label{fig:dd}
	\end{center}
	\vskip -0.3in
\end{figure*}

\begin{figure*}[!h]
	\vskip 0.2in
	\begin{center}
		\centering{\includegraphics[scale=0.8]{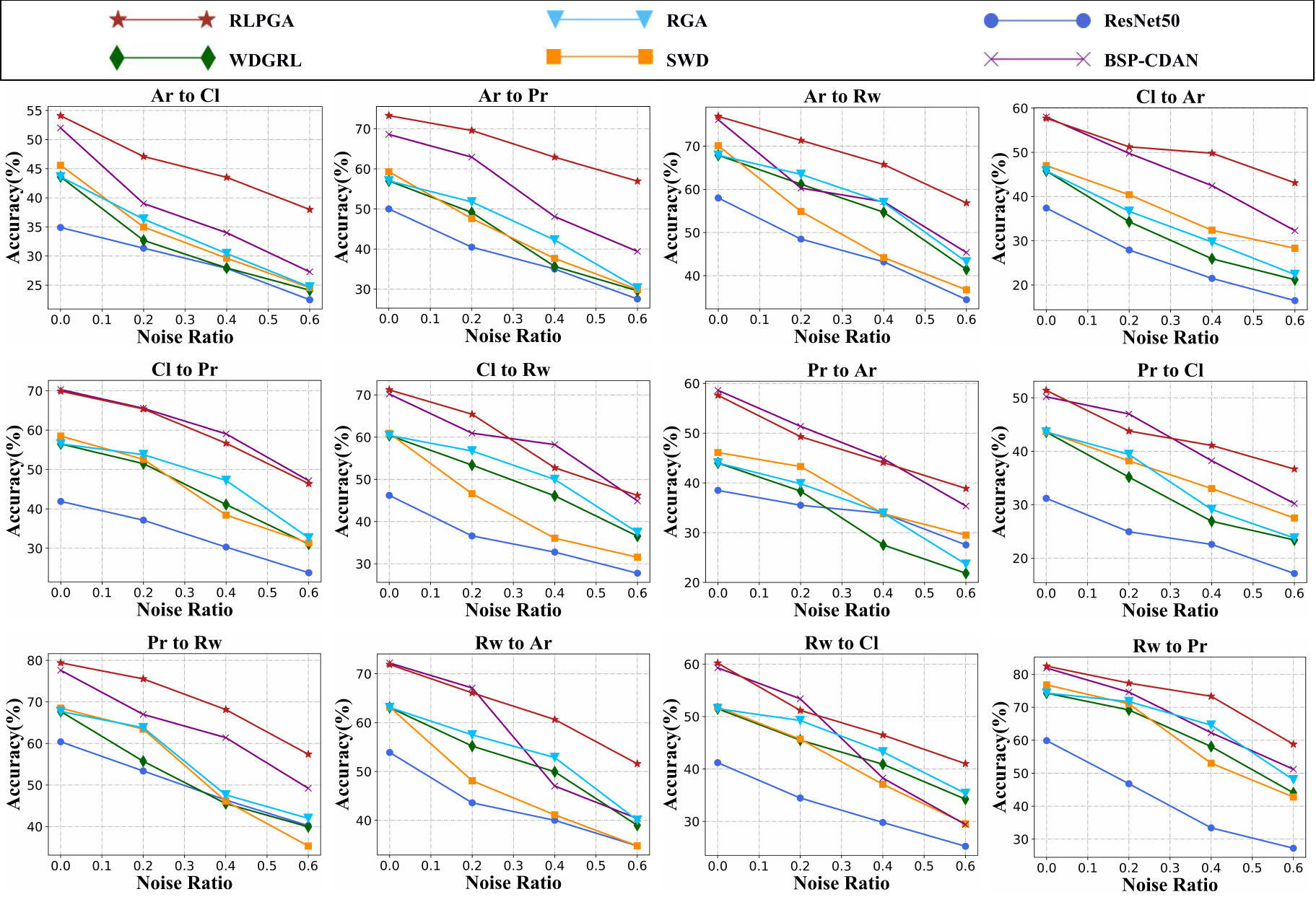}}
		\caption{Robustness evaluation on Office-Home dataset}
		\label{fig:ee}
	\end{center}
	\vskip -0.25in
\end{figure*}


\subsection{Denoising comparisons on transfer task} \label{dcomp}     
Fig. \ref{fig:bb}, \ref{fig:ee}, and \ref{fig:dd} show the experimental results when the labels of source domain samples are polluted by different noise ratios. The hyper-parameter $\alpha$ of RGA is also set to 0, which aims to eliminate the impact of the two weight graphs. We compare the performances of different methods based on specific transfer task classification accuracy and average classification accuracy.

\textit{1) Comparisons on the Office+Caltech10 dataset with DeCaf features:} From Fig. \ref{fig:bb},  we observe that RLPGA achieves the
best results under all noise ratios in 10 of 12 tasks. Especially,
the resulting curve of RLPGA is the most stable and least
decline with the addition of noise in all tasks. 

\textit{2) Comparisons on the Office31 dataset:} From Fig. \ref{fig:dd} (a), we
observe that RLPGA achieves the best results on 23 of 24
specific transfer tasks, and for A to D, A to W, D to W, W to A, and W to D transfer tasks, RLPGA achieves the best
results on all different noise ratios. Also, when the noise
ratio is equal to 0.6, the accuracy of RLPGA is at least 10\%
higher than the other five methods on average.

\textit{3) Comparisons on the Digits dataset:} From Fig. \ref{fig:dd} (b), we
observe that RLPGA achieves the best results on all specific
transfer tasks. Especially, when the noise ratio is equal to
0.6, the accuracy of RLPGA is almost 10\% higher than the
other five methods on average. Also, the curve of RLPGA is
smoother than other benchmark methods.

\textit{4) Comparisons on the Office-Home dataset:} From Fig. \ref{fig:ee}, we observe that RLPGA achieves the best results on 34 of 48
specific transfer tasks. For specific transfer tasks, such as Ar
to Cl, Ar to P r, Ar to Rw, P r to Rw, and Rw to P r, RLPGA
achieves the best results on 8 of 12 specific transfer tasks and
obtains comparable results on other 4 specific transfer tasks.
Also, when the noise ratio is equal to 0.6, the accuracy of
RLPGA is higher than the other five methods on all specific
transfer tasks.

Therefore, the denoising comparisons on the transfer task
can thoroughly verify the robustness against label noise of
our proposed RLPGA. 

\begin{figure*}[!h]
	\vskip 0.1in
	\begin{center}
		\centering{\includegraphics[width=0.95\textwidth]{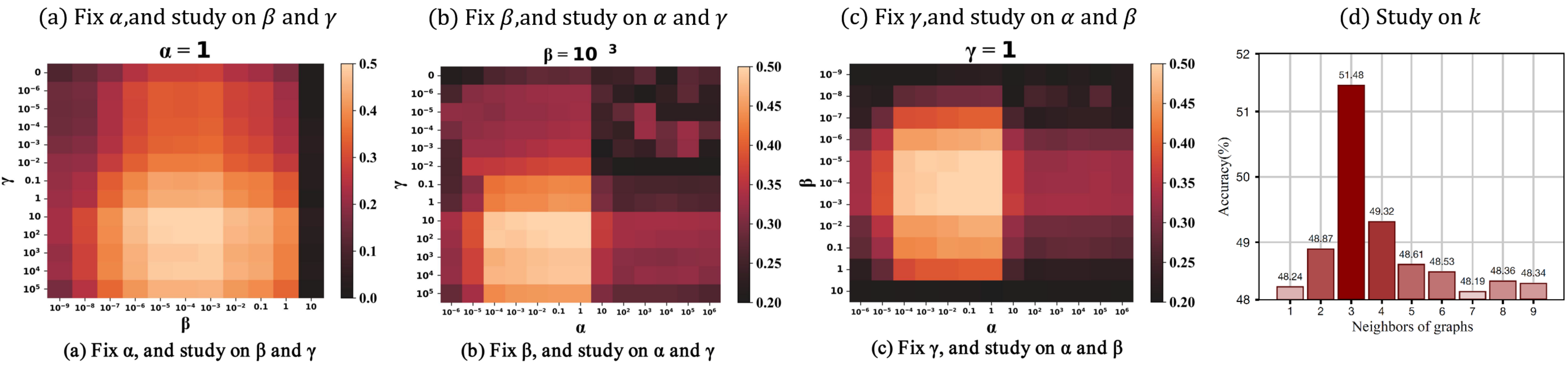}}
		\caption{The influence of hyper-parameters }
		\label{fig:heatmap}
	\end{center}
	\vskip -0.20in   	
\end{figure*}

\subsection{Ablation study} \label{as}
The proposed RLPGA is mainly composed of two parts
including the proposed robust informative theoretic-based
loss function and the constructed two adjacency weight matrices and two negative weight matrices to enhance the robustness to the noisy label. For ablation study,
a simplified version of RLPGA, which does not use the
two kind of weight matrices is verified and is named RGA. So, by
evaluating the classification accuracy of RGA on different
datasets under different noise ratio, we can verify whether the
robust informative theoretic-based loss function is effective
to improve the robustness. Comparing RLPGA with RGA,
we can evaluate whether the constructed four weight matrices
are effective to improve the robustness.

The main difference between RGA and WDGRL is that
RGA adopts the proposed robust informative theoretic-based loss function to improve the robustness to label
noise. Compared RGA with WDGRL, we observe that RGA
outperforms WDGRL in most specific transfer tasks under
different noise ratios of different data sets, where RGA
achieves better results in 31 tasks out of all 48 tasks on
the Amazon review dataset, 42 tasks out of all 48 tasks on
the Office-Caltech10 dataset with DeCaf features, 40 tasks
out of 48 tasks on the Office-Caltech10 dataset with SURF
features, 23 tasks out of 24 tasks on the Office31 dataset, 46
tasks out of 48 tasks on the Office-Home dataset, 10 tasks
out of 12 tasks on the Email Spam Filtering dataset, and 7
tasks out of 8 tasks on the Digits datasets. Therefore, we
can conclude that it is effective to consider the proposed
robust informative theoretic-based loss function to reduce
the sensitivity to label noise.

The main difference between RLPGA and RGA is that
RLPGA constructs two weight graphs to enhance the feature
discriminability, thereby reducing the influence of noise
labels on the learned feature representation. We observe
that RLPGA obtains better classification accuracy in many
specific transfer tasks on different datasets, where RLPGA
achieves better results in 41 tasks out of all 48 tasks on the
Amazon review dataset, 44 tasks out of all 48 tasks on the
Office-Caltech10 dataset with DeCaf features, 40 tasks out of
48 tasks on the Office-Caltech10 dataset with SURF features,
23 tasks out of 24 tasks on the Office31 dataset, all tasks
on the Office-Home dataset, 11 tasks out of 12 tasks on the
Email Spam Filtering dataset, and all tasks on the Digits
datasets. Therefore, we can conclude that it is effective to
reduce the influence of noise labels on the learned feature
representation by enhancing the feature discriminability.

Note that many methods focus on adaptation architecture design. As for our proposed method, we mainly focus
on the objective function, and the new network structures
can be easily integrated into our framework. Even so, from
the discussion in Subsection \ref{sec:implement}, we can know that the
average accuracy of RLPGA outperforms most compared
methods on most datasets. Therefore, we can conclude that
both the proposed robust informative theoretic-based loss and the constructed two weight graphs are effective to
improve the robustness to label noise.


\begin{figure*}[!h]
	\vskip 0.005in
	\begin{center}
		\centering{\includegraphics[scale=0.80]{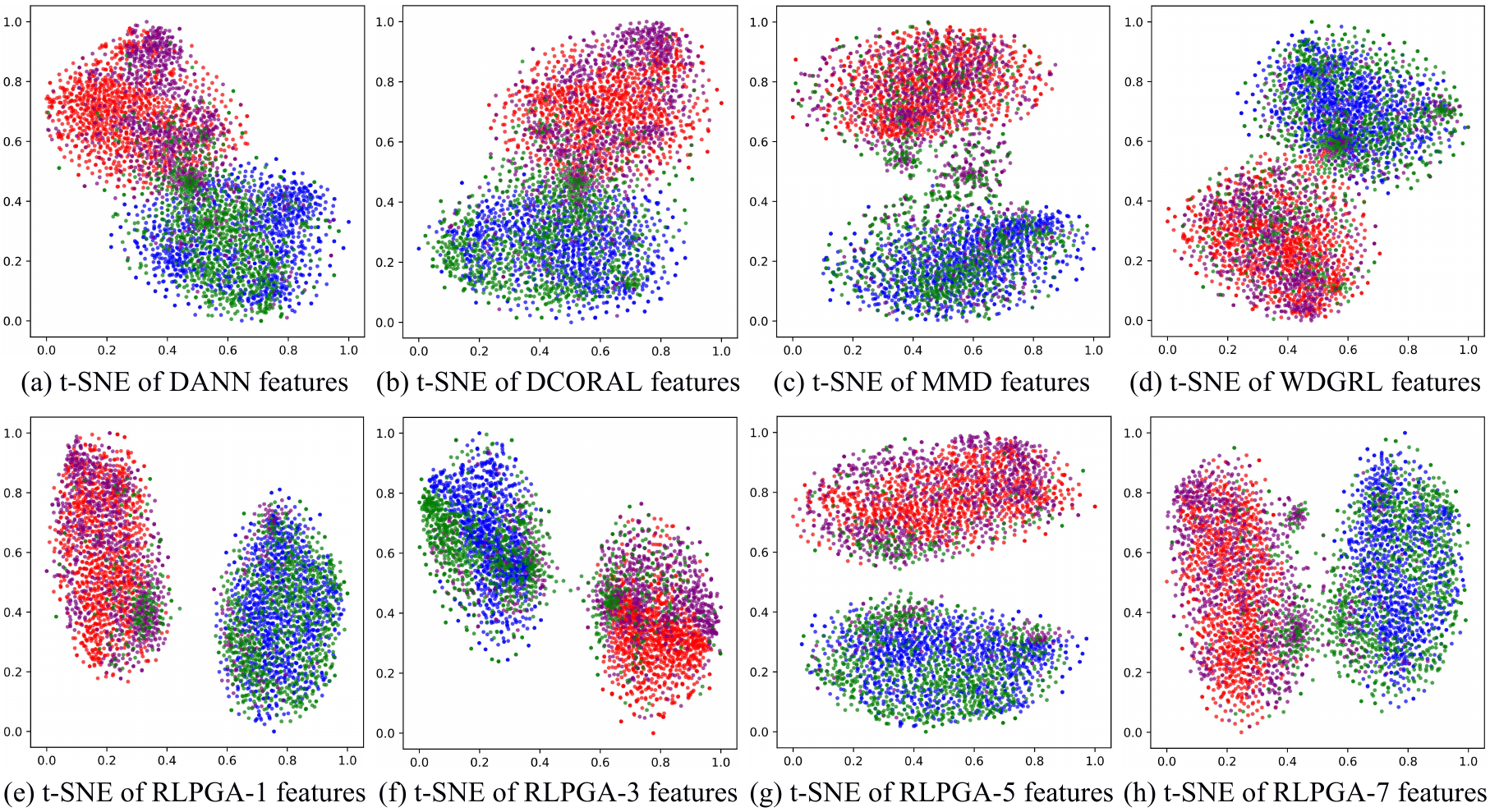}}
		\caption{Feature visualization of the $K \to E$ task in Amazon review dataset}
		\label{fig:ffff}
	\end{center}
	\vskip -0.2in
\end{figure*}\label{key}

\subsection{Influence of Hyper-parameters} \label{tf}
Specifically, we performed several experiments to study the influence of the hyper-parameters in our proposed RLPGA
including $\alpha$ which is used to balance the impact of the term $Di{s_{pn}}\left( f \right)$, $\beta$ which is used to balance the impact of the term ${W_p}\left( {P_s^{f\left( X \right)},P_t^{f\left( X \right)}} \right)$, $\gamma$ which is used to balance the impact of the term $L_{r}$. To intuitively understand the influences of the hyper-parameters, we take several experiments based on the transfer task Pr$\to$Cl of Office-Home dataset. As the results are
shown in Fig. \ref{fig:heatmap}, the plots further elaborate our deepgoing
studies’ results with RLPGA. To explore the influence of $\alpha$, we first fix $\beta={10^{3}}$, $\gamma=1$ and $k=3$ and then select the $\alpha$ from range of $\left\{ {{\rm{1}}{{\rm{0}}^{{\rm{ - 6}}}},...,{\rm{1}}{{\rm{0}}^{{\rm{ - 1}}}}{\rm{,1,1}}{{\rm{0}}^{\rm{1}}},...,{\rm{1}}{{\rm{0}}^{\rm{6}}}} \right\}$. From the results, we observe that appropriate enhancement of feature discrimination can promote the performance of our proposed method. To explore the influence of $\beta$, we first fix $\alpha=1$, $\gamma=1$ and $k=3$ and then select the $\beta$ from range of $\left\{ {{\rm{1}}{{\rm{0}}^{ - 9}},...,{{10}^3}} \right\}$. From Fig. \ref{fig:heatmap}, we observe that the transferability of learned feature representation is important to the classification task. To explore the influence of $\gamma$, we first fix $\alpha={1}$, $\beta={{10}^3}$, $k=3$, and then select the $\gamma$ from range of $\left\{ {0,{{10}^{ - 5}},...,1,...,{{10}^5}} \right\}$. From the results, we observe that the cross entropy loss and $L_{r}$ are all important to the classification task. 

We further conduct experiments based on the transfer task Pr$\to$Cl of Office-Home dataset to explore the influence of the number of neighbor points $k$, and the number of neighbor points $k$ when constructing two weight matrices. We first fix $\alpha={1}$, $\beta={{10}^3}$, $\gamma=1$, and then select the $k$ from range of $\left\{ {1,...,9} \right\}$. The results are shown in Fig. \ref{fig:heatmap} (d), we can see that an appropriate number of neighbor points is important.  

\begin{figure*}[!h]
	\vskip 0.1in
	\begin{center}
		\centering{\includegraphics[scale=0.5]{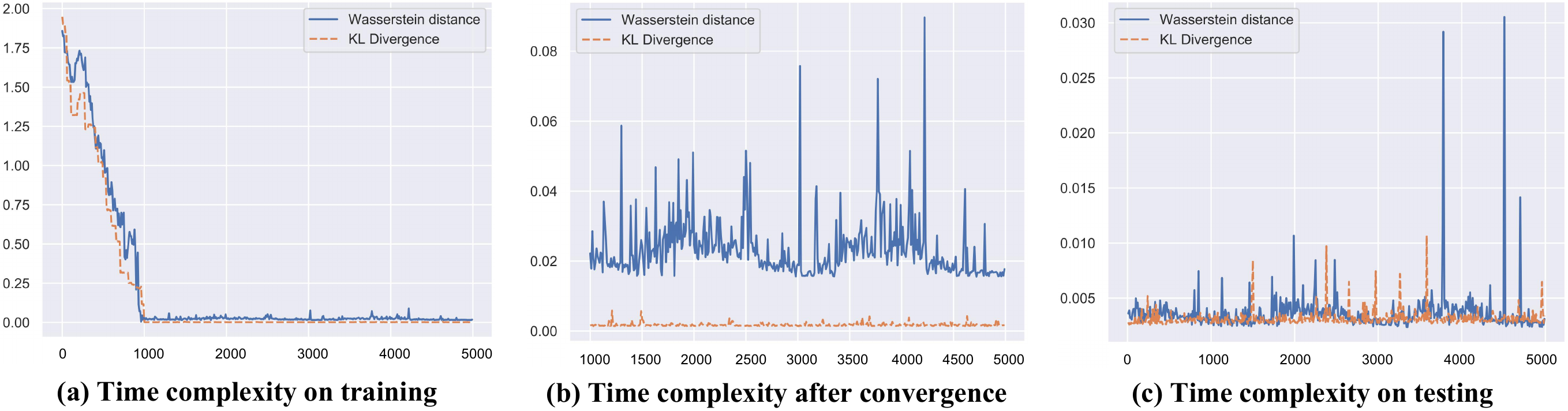}}
		\caption{The analysis of time complexity on different discrepancy metrics}
		\label{timeanalysisa}
	\end{center}
	\vskip -0.25in   	
\end{figure*}

\begin{figure*}[!h]
	\vskip 0.1in
	\begin{center}
		\centering{\includegraphics[scale=0.5]{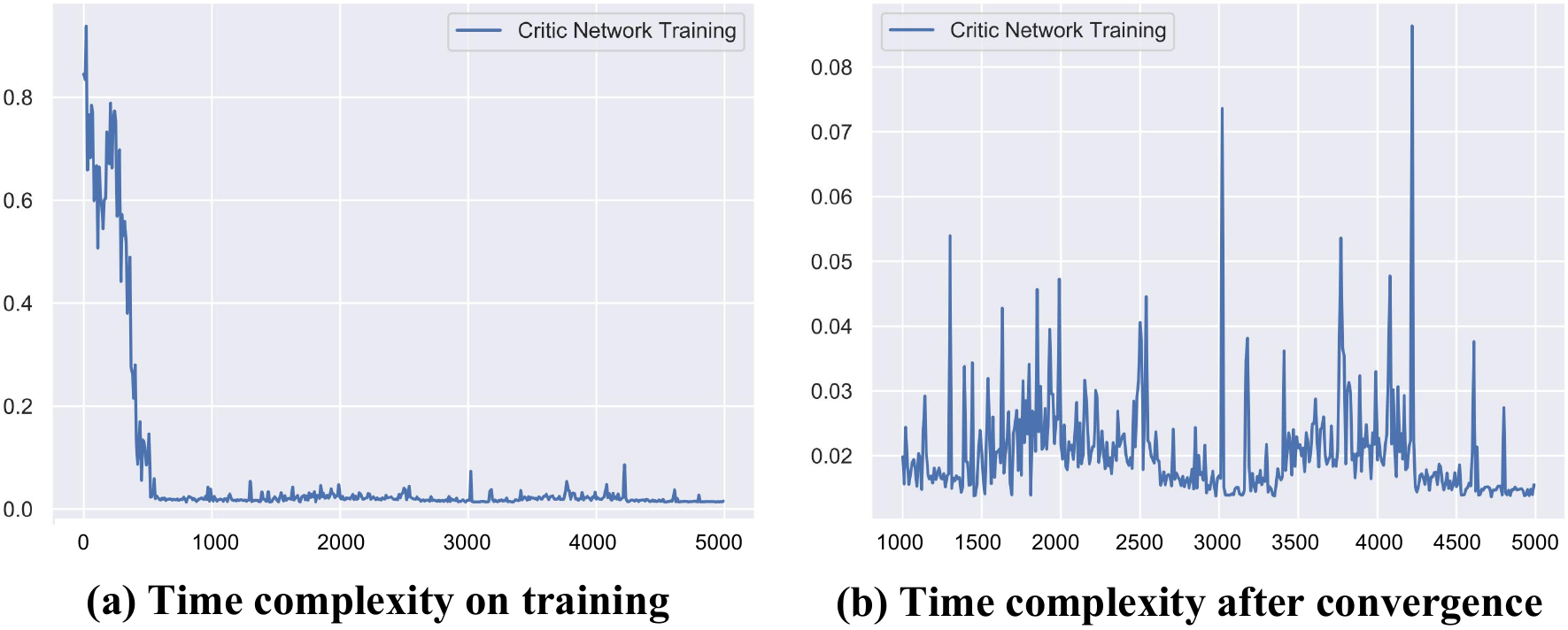}}
		\caption{The analysis of time complexity on the training stage of critic network}
		\label{timeanalysisb}
	\end{center}
	\vskip -0.25in   	
\end{figure*}

\subsection{The deepgoing analysis of the discrepancy metric}
The time complexity of our proposed method will vary depending on the specific discrepancy metric taken, for instance, Wasserstein distance, KL divergence, etc. For the exact purpose of exploring the time complexity of RLPGA based on Wasserstein distance or KL divergence, we conduct several comparisons on the synthetic dataset. Fig. \ref{timeanalysisa} shows the experimental results of time complexity, and in details, (a) represents the time complexity stats on the training stage of all time, (b) denotes the results on training stage after the convergence, and (c) represents the results on the testing stage of all time. The figure (a) actively demonstrates that in general, the time complexity of RLPGA with Wasserstein distance (i.e., RLPGA w/ WD) is higher than that of RLPGA with KL divergence (i.e., RLPGA w/ KL), but the difference is not significantly large. From figure (b), we observe that after convergence, the complexity of RLPGA w/ WD is more unstable than RLPGA w/ KL. As shown in figure (c), in the testing stage, the complexities of the compared methods are similar, to some degree.

Fig. \ref{timeanalysisb} demonstrates the time complexity stats of the critic network on the training stage, and the subfigure (a) represents the records of all time, and the subfigure (b) denotes the records after convergence. From figure (a) of Fig. \ref{timeanalysisa} and figure (a) of Fig. \ref{timeanalysisb}, we observe that compared with the main model of our proposed RLPGA, the critic network of the discrepancy metric reaches the convergence state faster. In addition, by observing Fig. \ref{timeanalysisa} (b) and Fig. \ref{timeanalysisb} (b), we find that after convergence, the fluctuation of time complexity is mainly brought by the critic network, and the particular reason for this circumstance is that with the entry of new batch of data, the critic network for the distribution discrepancy calculation of the source domain and the target domain will be updated all the way. However, after convergence, with the entry of a new batch of data, the main model of RLPGA will generate slight gradients, and the fluctuation of the optimization is accordingly trivial.

\begin{table}[!t]
	\caption{Performance (accuracy) on Digits dataset with different discrepancy metrics}
	\label{tab:dmcomp}
	\setlength{\tabcolsep}{1.9pt}
	\vskip 0.01in
	\begin{center}
		\begin{small}
			\begin{tabular}{c|ccc}
				\hline
				Domains & MNIST$\to$USPS & USPS$\to$MNIST & Average \\
				\hline
				\hline
				DANN & 90.4 & 94.7 & 92.6 \\
				ADDA & 89.4 & 90.1 & 89.8 \\
				UNIT & 96.0 & 93.6 & 94.8 \\
				CyCADA & 95.6 & 96.5 & 96.1 \\
				CDAN & 93.9 & 96.9 & 95.4 \\
				CDAN+E & 95.6 & 98.0 & 96.8 \\
				BSP+DANN & 94.5 & 97.7 & 96.1 \\
				BSP+ADDA & 93.3 & 94.5 & 93.9 \\
				BSP+CDAN & 95.0 & 98.1 & 96.6 \\
				SWD & \textbf{98.1} & 97.1 & 97.6 \\
				\hline
				\hline
				RLPGA w/ KL & 93.6 & 95.3 & 94.5 \\
				\textbf{RLPGA w/ WD} & 97.2 & \textbf{98.6} & \textbf{97.9} \\
				\hline
			\end{tabular}
		\end{small}
	\end{center}
	\vskip -0.1in
\end{table}

Along the lines of the experimental principle of Section \ref{sec:ccomp}, we further conduct experiments on the Digits dataset to clarify the performance of RLPGA w/ WD and RLPGA w/ KL. As demonstrated in Table \ref{tab:dmcomp}, we reckon that although the calculation of Wasserstein distance is more time-consuming than that of KL divergence, the former discrepancy metric can better depict the differences between the distributions of the source domain and the target domain. In detail, RLPGA w/ WD outperforms RLPGA w/ KL by 2.6\% on MNIST$\to$USPS task and 3.3\% on USPS$\to$MNIST task. Therefore, we adopt Wasserstein distance as the specific discrepancy metric for the proposed RLPGA.

\subsection{Feature Visualization} \label{fv}
To show the feature transferability and discriminability intuitively, we set the noise ratio $r$ as 0.2 and visualize the features learned by the eight methods based on the $K \to E$ transfer task of Amazon review dataset. We introduce the t-SNE visualization to visualize the learned features and plot them in Fig. \ref{fig:ffff}. For all subgraphs in Fig. \ref{fig:ffff}, red and blue dots separately represent positive and negative samples in the source domain, and purple and green dots represent positive and negative samples in the target domain, respectively. High feature transferability should bring together dots of the same class in both domains, while high feature discriminability should separate dots of different classes from each other. We observe the feature transferability is learned well for all approaches. As for the feature discriminability, the representations learned by RLPGA outperform other approaches. So, this indicates that the proposed RLPGA is more effective and robust.

\section{Conclusions}
In this paper, we propose a novel method called robust local preserving and global aligning network for adversarial domain adaptation (RLPGA). RLPGA tackles the problem of learning domain adaptation models under the setting of noisy labels. First, RLPGA introduces a robust loss for solving this problem. We prove that it can reduce the impact of label noises. Then, to reduce the effect of label noises from the feature perspective, a local preserving and global aligning method is proposed. We also provide a theoretical analysis that RLPGA is conducive to minimize the target risk. Experiments results on sentiment and image classification domain adaptation datasets show the effectiveness of the proposed method.
classification of RLPGA is 83.4\%, which is 1\% higher than the best among the 5 domain adaptation methods. For specific transfer task classification accuracy, RLPGA achieves the best results on 12 specific transfer tasks. The improvements of our Proposed RLPGA in this dataset are significant. 

\section{Acknowledgements}
The authors would like to thank the associate editor and anonymous reviewers for their valuable comments. This work is supported in part by National Natural Science Foundation of China No. 61976206, No. 61832017, No. 91746301, No. 71531001, Beijing Outstanding Young Scientist Program NO. BJJWZYJH012019100020098, CCF-Tencent Open Fund RAGR20200110, the Fundamental Research Funds for the Central Universities, the Research Funds of Renmin University of China 21XNLG05, Public Computing Cloud, Renmin University of China, Key Special Project for Introduced Talents Team of Southern Marine Science and Engineering Guangdong Laboratory (Guangzhou) No. GML2019ZD0603, and Strategic Priority Research Program of the Chinese Academy of Sciences Grant No. XDA19020500. This work is also supported in part by Intelligent Social Governance Platform, Major Innovation \& Planning Interdisciplinary Platform for the “Double-First Class” Initiative, Renmin University of China, and Public Policy and Decision-making Research Lab of Renmin University of China.

\ifCLASSOPTIONcaptionsoff
\newpage
\fi



\bibliographystyle{IEEEtran}
\bibliography{mybibfile}
\clearpage
%



%
\vskip -0.5in
\begin{IEEEbiography}[{\includegraphics[width=1in,height=1.25in,clip,keepaspectratio]{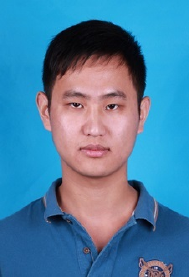}}]{Wenwen Qiang}
	received the MS degree in the department of mathematics, college of science, China Agricultural University, Beijing, in 2018. He is currently a doctoral student at the University of Chinese Academy of Sciences. His research interests include transfer learning, deep learning, and machine learning.
\end{IEEEbiography}
\vskip -0.25in
\begin{IEEEbiography}[{\includegraphics[width=1in,height=1.25in,clip,keepaspectratio]{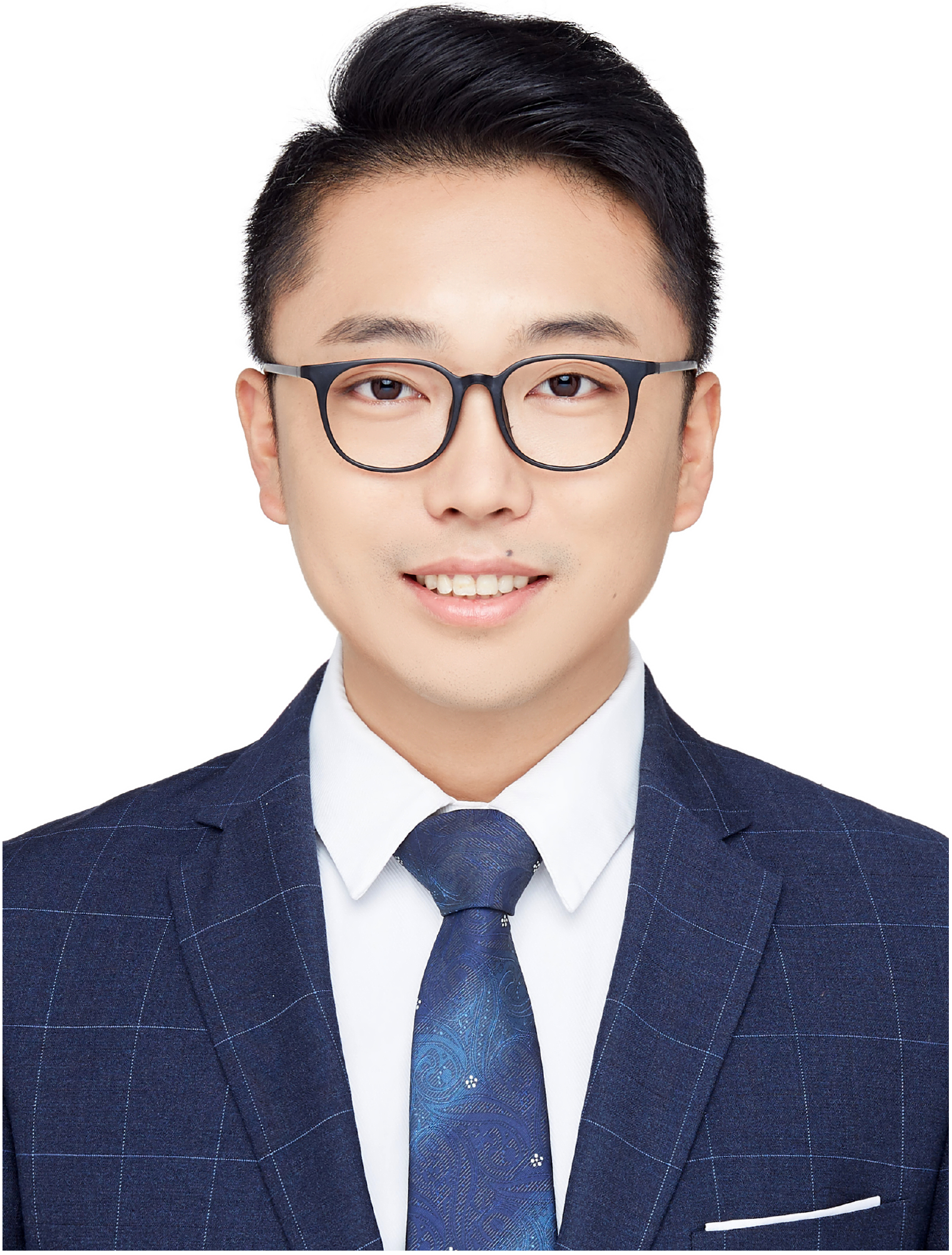}}]{Jiangmeng Li}
	received the MS degree with concentration of data science, School of Professional Studies, New York University, New York, New York, USA, in 2018. He is currently a doctoral student at the University of Chinese Academy of Sciences. His research interests include transfer learning, deep learning, and machine learning.
\end{IEEEbiography}
\vskip -0.25in
\begin{IEEEbiography}[{\includegraphics[width=1in,height=1.25in,clip,keepaspectratio]{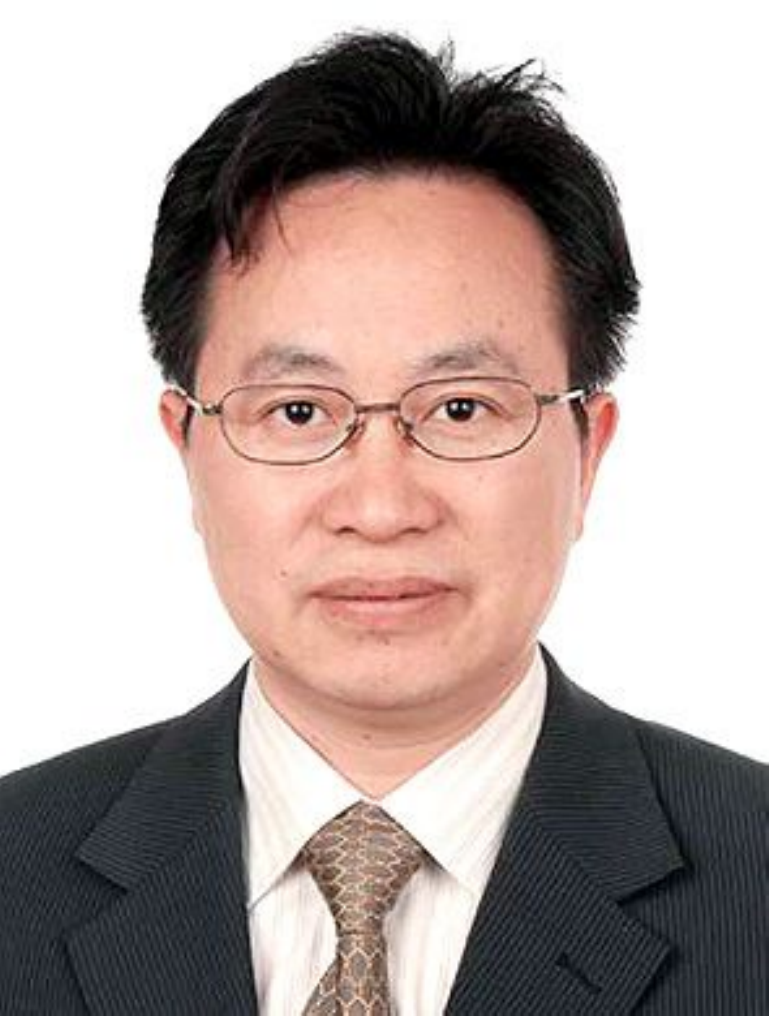}}]{Changwen Zhen}
	received the Ph.D. degree in Huazhong University of Science and Technology. He is currently a professor in Institute of Software, Chinese Academy of Science. His research interests include computer graph and artificial intelligence.
\end{IEEEbiography}
\vskip -0.25in


\begin{IEEEbiography}[{\includegraphics[width=1in,height=1.25in,clip,keepaspectratio]{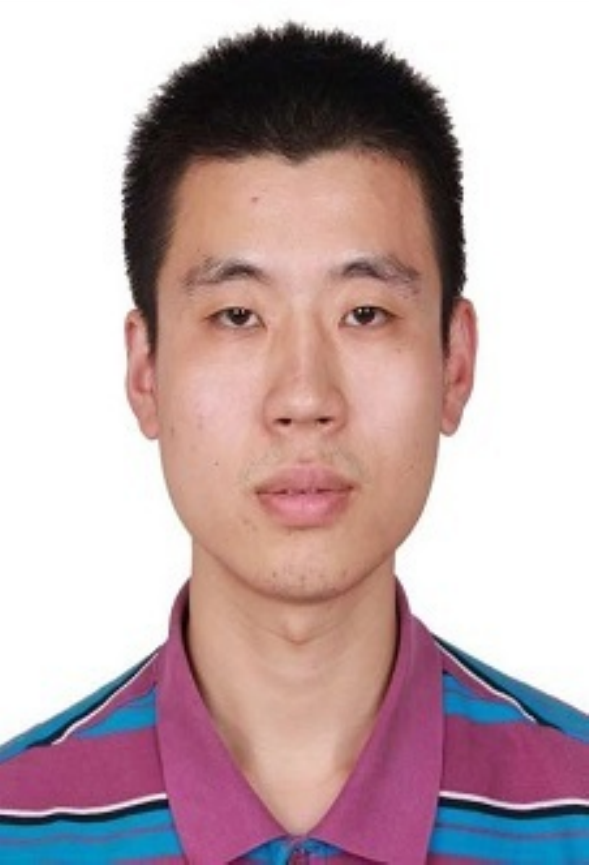}}]{Bing Su}
	received the BS degree in information engineering from the Beijing Institute of Technology, Beijing, China, in 2010, and the PhD degree in electronic engineering from Tsinghua University, Beijing, China, in 2016. From 2016 to 2020, he worked with the Institute of Software, Chinese Academy of Sciences, Beijing. Currently, he is an associate professor with the Gaoling School of Artificial Intelligence, Renmin University of China. His research interests include pattern recognition, computer vision, and machine learning. He has published more than ten papers in journals and conferences such as IEEE Transactions on Pattern Analysis and Machine Intelligence (TPAMI), IEEE Transactions on Image Processing (TIP), International Conference on Machine Learning (ICML), IEEE Conference on Computer Vision and Pattern Recognition (CVPR), IEEE International Conference on Computer Vision (ICCV), etc.
\end{IEEEbiography}
\vskip -0.35in

\begin{IEEEbiography}[{\includegraphics[width=1in,height=1.25in,clip,keepaspectratio]{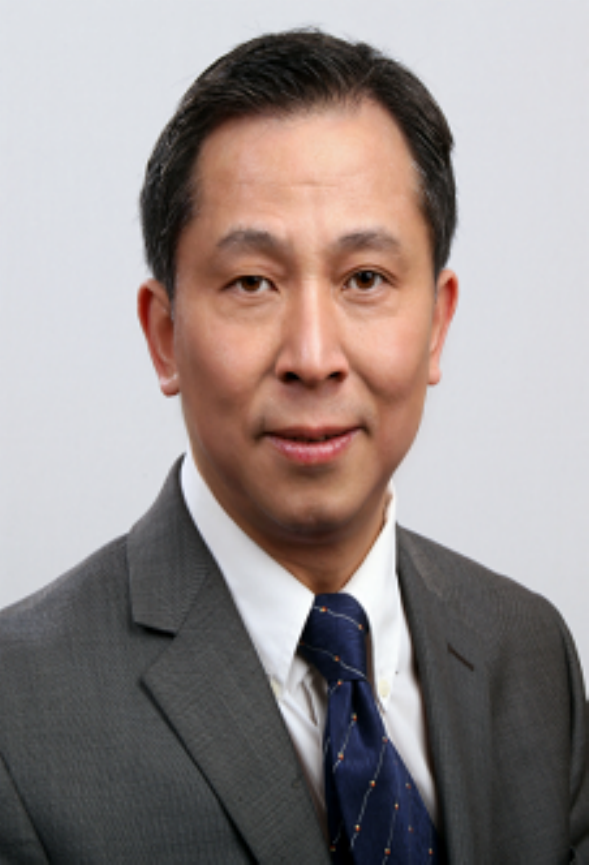}}]{Hui Xiong}
	received his Ph.D. in Computer Science from the University of Minnesota - Twin Cities, USA, in 2005, the B.E. degree in Automation from the University of Science and Technology of China (USTC), Hefei, China, and the M.S. degree in Computer Science from the National University of Singapore (NUS), Singapore. He is a chair professor at the Hong Kong University of Science and Technology (Guangzhou). He is also a Distinguished Professor at Rutgers, the State University of New Jersey, where he received the 2018 Ram Charan Management Practice Award as the Grand Prix winner from the Harvard Business Review, RBS Dean's Research Professorship (2016), two-year early promotion/tenure (2009), the Rutgers University Board of Trustees Research Fellowship for Scholarly Excellence (2009), the ICDM-2011 Best Research Paper Award (2011), the Junior Faculty Teaching Excellence Award (2007), Dean's Award for Meritorious Research (2010, 2011, 2013, 2015) at Rutgers Business School, the 2017 IEEE ICDM Outstanding Service Award (2017), and the AAAI-2021 Best Paper Award (2021). Dr. Xiong is also a Distinguished Guest Professor (Grand Master Chair Professor) at the University of Science and Technology of China (USTC). For his outstanding contributions to data mining and mobile computing, he was elected an ACM Distinguished Scientist in 2014, an IEEE Fellow and an AAAS Fellow in 2020. His general area of research is data and knowledge engineering, with a focus on developing effective and efficient data analysis techniques for emerging data intensive applications. He currently serves as a co-Editor-in-Chief of Encyclopedia of GIS (Springer) and an Associate Editor of IEEE Transactions on Data and Knowledge Engineering (TKDE), IEEE Transactions on Big Data (TBD), ACM Transactions on Knowledge Discovery from Data (TKDD) and ACM Transactions on Management Information Systems (TMIS). He has served regularly on the organization and program committees of numerous conferences, including as a Program Co-Chair of the Industrial and Government Track for the 18th ACM SIGKDD International Conference on Knowledge Discovery and Data Mining (KDD), a Program Co-Chair for the IEEE 2013 International Conference on Data Mining (ICDM), a General Co-Chair for the IEEE 2015 International Conference on Data Mining (ICDM), and a Program Co-Chair of the Research Track for the 24th ACM SIGKDD International Conference on Knowledge Discovery and Data Mining (KDD2018).
	
\end{IEEEbiography}




\clearpage

\section{Appendix}\label{sec:appendix}

\begin{table}[!h]
	\caption{Performance (accuracy) on Email dataset}
	\label{tab:ff}
	\setlength{\tabcolsep}{3.5pt}
	\vskip 0.1in
	\begin{center}
		\begin{small}
			\begin{tabular}{c|cccc}
				\hline
				Domains & P $ \to $ ${u_1}$  & P $ \to $ ${u_2}$  & P $ \to $ ${u_3}$  & Average \\
				\hline
				\hline
				MMD & 81.0 & 86.0 & 94.1 & 87.0 \\
				DANN & 83.3 & 85.7 & 91.9 & 86.9 \\
				DCORAL & 79.7 & 83.8 & 89.8 & 84.4 \\
				WDGRL & 85.7 & 88.3 & 95.8 & 89.9 \\
				SWD & 87.2 & 88.8 & 94.5 & 90.2 \\
				\hline
				\hline
				\textbf{RLPGA} & \textbf{87.6} & \textbf{89.1} & \textbf{97.1} & \textbf{91.3} \\
				\hline
			\end{tabular}
			
		\end{small}
	\end{center}
	\vskip -0.2in
\end{table}

\begin{table*}[!h]
	\caption{Performance (accuracy) on Amazon review dataset}
	\label{tab:aa}
	\setlength{\tabcolsep}{2.6pt}
	\vskip 0.1in
	\begin{center}
		\begin{small}
			\begin{tabular}{c|ccccccccccccc}
				\hline
				Domains& B$\to$D & B$\to$E & B$\to$K & D$\to$B & D$\to$E & D$\to$K & E$\to$B & E$\to$D & E$\to$K & K$\to$B & K$\to$D & K$\to$E & Average \\
				\hline
				\hline
				MMD& 82.6 & 80.9 & 83.5 & 79.9 & 82.5 & 84.1 & 75.7 & 77.7 & 87.4 & 75.8 & 78.1 & 86.3 & 81.2 \\
				DANN& 82.1 & 78.9 & 82.7 & 79.3 & 81.6 & 83.4 & 75.9 & 77.6 & 86.6 & 75.8 & 78.5 & 86.1 & 80.7 \\
				DCORAL& 82.7 & 82.9 & 84.8 & 80.8 & 83.4 & 85.3 & 76.9 & 78.1 & 87.9 & 76.9 & 79.1 & 86.8 & 82.2 \\
				WDGRL& 83.1 & 83.2 & 85.4 & 80.7 & 83.5 & 86.2 & 77.2 & 78.3 & 88.2 & 77.2 & 79.9 & 86.3 & 82.4 \\
				SWD& 82.9 & 83.1 & 85.1 & 80.5 & 83.7 & 85.9 & 77.1 & 78.6 & 87.7 & 76.6 & 79.6 & 86.1 & 82.2 \\
				\hline
				\hline
				\textbf{RLPGA}& \textbf{83.7} & \textbf{83.9} & \textbf{85.7} & \textbf{81.9} & \textbf{84.2} & \textbf{87.0} & \textbf{78.8} & \textbf{80.5} & \textbf{88.6} & \textbf{78.3} & \textbf{80.8} & \textbf{87.1} & \textbf{83.4}\\
				\hline
			\end{tabular}
		\end{small}
	\end{center}
	\vskip -0.2in
\end{table*}

\begin{table*}[!h]
	\caption{Performance (accuracy) on Office+Caltech10 dataset with SURF features}
	\label{tab:cc}
	\setlength{\tabcolsep}{1.9pt}
	\vskip 0.1in
	\begin{center}
		\begin{small}
			\begin{tabular}{c|ccccccccccccc}
				\hline
				Domains & A$\to$C & A$\to$D & A$\to$W & W$\to$A & W$\to$D & W$\to$C & D$\to$A & D$\to$W & D$\to$C & C$\to$A & C$\to$W & C$\to$D & Average \\
				\hline
				\hline
				TJM & 39.5 & 45.2 & 42.0 & 30.0 & 89.2 & 30.2 & 32.8 & 85.4 & 31.4 & 46.8 & 39.0 & 44.6 & 46.3 \\
				SCA & 39.7 & 39.5 & 34.9 & 30 & 87.3 & 31.1 & 31.6 & 84.4 & 30.7 & 45.6 & 40.0 & 47.1 & 45.2 \\
				ARTL & 36.1 & 36.9 & 33.6 & 38.3 & 87.9 & 29.7 & 34.9 & 88.5 & 30.5 & 44.1 & 31.5 & 39.5 & 44.3 \\
				JGSA & 41.5 & 47.1 & 45.8 & 39.9 & \textbf{90.5} & 33.2 & 38.0 & \textbf{91.9} & 29.9 & 51.5 & 45.4 & 45.9 & 50.0 \\
				CORAL & 45.1 & 39.5 & 44.4 & 36.0 & 86.6 & 33.7 & 37.7 & 84.7 & 33.8 & 52.1 & 46.4 & 45.9 & 48.8 \\
				\hline
				\hline
				MMD & 44.1 & 41.4 & 37.3 & 34.1 & 84.7 & 30.7 & 32.5 & 73.6 & 30.7 & 54.8 & 40.3 & 47.1 & 45.9 \\
				DANN & 45.0 & 41.4 & 38.6 & 34.1 & 82.8 & 32.7 & 31.6 & 74.2 & 32.2 & 54.9 & 43.4 & 47.8 & 46.6\\
				DCORAL & 45.0 & 40.1 & 38.3 & 34.9 & 84.1 & 33.3 & 31.5 & 73.9 & 31.5 & 53.4 & 40.0 & 47.1 & 46.1\\
				WDGRL & 45.9 & 44.6 & 40.7 & 32.2 & 81.5 & 31.1 & 35.6 & 77.0 & 32.6 & 55.2 & 42.4 & 48.4 & 47.3 \\
				SWD & 44.1 & 42.3 & 40.5 & 32.2 & 80.4 & 30.6 & 32.9 & 75.6 & 33.1 & 54.8 & 41.9 & 48.9 & 46.4 \\
				MEDA & 43.9 & 45.9 & \textbf{53.2} & \textbf{42.7} & 88.5 & \textbf{34.0} & 41.2 & 87.5 & \textbf{34.9} & 56.5 & 53.9 & 50.3 & 52.7 \\
				\hline
				\hline
				\textbf{RLPGA} & \textbf{54.5} & \textbf{52.6} & 46.3 & 36.7 & 84.2 & 30.1 & \textbf{42.4} & 77.9 & 32.5 & \textbf{62.7} & \textbf{54.7} & \textbf{66.7} & \textbf{53.4} \\
				\hline
			\end{tabular}
		\end{small}
	\end{center}
	\vskip -0.2in
\end{table*}

\subsection{Datesets} 
\textbf{Amazon review dataset} \cite{bli01} records the product reviews on Amazon.com and includes four domains, \textit{e.g.,} books (B), DVDs (D), electronics (E), and kitchen appliances (K). \textbf{Office-Caltech10 dataset} \cite{gong201} includes four domains, \textit{e.g.,} Amazon (A), Webcam (W), DSLR (D). and Caltech (C). We adopt 800-dimensional SURF feature for the samples in Office+Caltech10 dataset. \textbf{Email Spam Filtering dataset} \cite{saenko2010} contains four user inboxes. We set the public inbox as the source domain and the other three private inboxes as target domains.

\subsection{Extended conventional comparisons} 
\label{sec:extendccomp}

We conducted extended conventional unsupervised domain
adaptation experiments of transferring task on benchmark
datasets, and the reported tables show the experimental
results when the source domain samples have ground true
labels. Note that only 5 baselines are presented in Table \ref{tab:ff} and Table \ref{tab:aa}. The first reason is that all experimental results of the compared methods in our submission are quoted from their respective original papers. To ensure the fairness and authenticity of the experimental results, we have not reproduced the experimental results of the compared methods on the data set that did not appear in the original article. The second reason is that most of the compared methods are based on image data, the Amazon dataset and Email dataset are not an image dataset.

\textit{1) Comparisons on the Email Spam Filtering dataset:} From Table \ref{tab:ff}, we observe that the best result always appears in the last column. RLPGA achieves the best average classification
and reaches 91.3\%, which is 1.4\% higher than the best
among the compared 4 domain adaptation methods and
7.9\% higher than DCORAL. Also, RLPGA achieves the best
classification accuracies on all specific transfer tasks. We
can also observe that adversarial based methods such as
RLPGA, WDGRL, and DANN are better than MMD and
DCORAL. The improvements of our Proposed RLPGA in this dataset are significant.

\textit{2) Comparisons on the Amazon review dataset:} From Table \ref{tab:aa}, we observe that the best result always appears in the last line. For specific transfer tasks, RLPGA achieves the best results on 12 specific transfer tasks. E.g., the average
classification of RLPGA is 83.4\%, which is 1\% higher than the best among the 5 domain adaptation methods. For specific transfer task classification accuracy, RLPGA achieves the best results on 12 specific transfer tasks. The improvements of our Proposed RLPGA in this dataset are significant. 

\textit{3) Comparisons on the Office+Caltech10 dataset with SURF features:} From Table \ref{tab:cc}, we observe that the proposed RLPGA has achieved the best results in more than half of the tasks. For specific transfer tasks, RLPGA achieves the best results on 6 of 12 specific transfer tasks and is the one that has achieved the best results the most times, e.g., the average classification of RLPGA is 53.4\%, which is 0.7\% higher than the best result among the other 14 domain adaptation methods, 6.1\% higher than WDGRL and, 7.3\% higher than DCORAL. As for specific transfer task classification accuracy, RLPGA achieves the best results on 6 of 12 specific transfer tasks and is the one that has achieved the best results the most times. Also, we can conclude that the results of deep domain adaptation methods are overall better than the results of traditional learning methods. The improvements of our Proposed RLPGA in this dataset are significant.

\subsection{Extended denoising comparisons} 
\label{sec:extenddcomp}

The figures show the extended experimental results when the labels of source domain samples are polluted by different noise ratios.

\begin{figure*}[!h]
	\vskip 0.2in
	\begin{center}
		\centering{\includegraphics[scale=0.66]{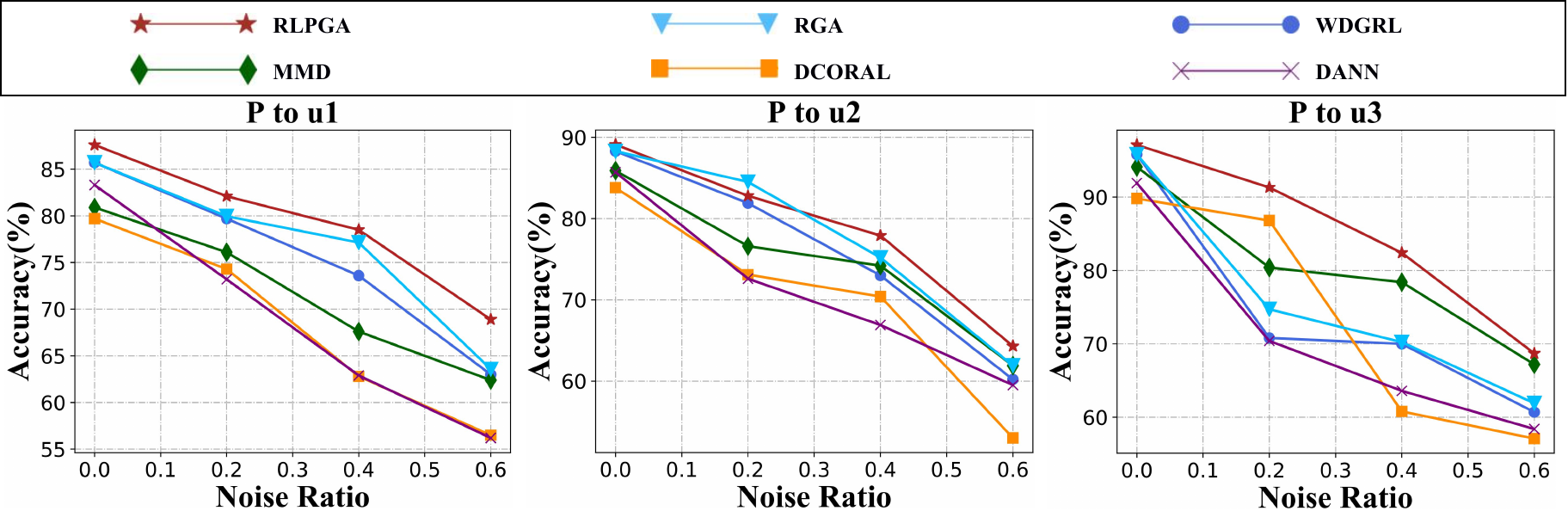}}
		\caption{Robustness evaluation on Email Spam Filtering dataset}
		\label{fig:ff}
	\end{center}
	\vskip -0.2in
\end{figure*}

\textit{1) Comparisons on the Email Spam Filtering dataset:} From Fig. \ref{fig:ff}, we observe that RLPGA achieves the best results
on 10 of 12 specific transfer tasks. Especially, for P to u1 and
P to u3 transfer tasks, RLPGA achieves the best results on
all different noise ratios. Also, the curve of RLPGA has the
least decline.

\begin{figure*}[!h]
	\vskip 0.1in
	\centering		
	\includegraphics[scale=0.8]{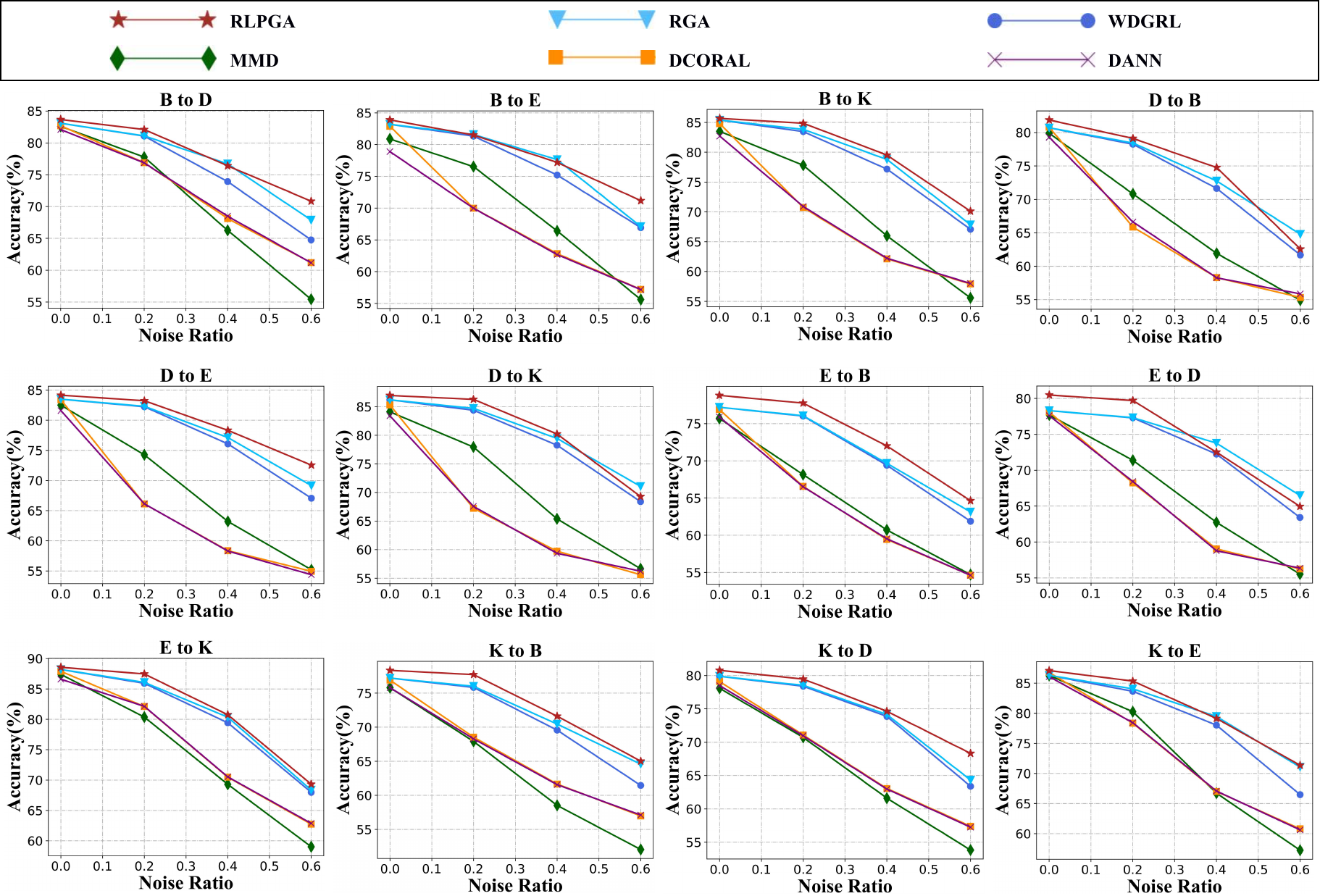}
	\captionsetup{justification=centering}
	\caption{Robustness evaluation on Amazon review dataset}
	\label{fig:aa}			
	\vskip -0.2in	
\end{figure*}

\textit{2) Comparisons on the Amazon review dataset:} Fig. \ref{fig:aa} shows the experimental results with different noise ratios.
Compared RLPGA with the other 4 methods, we observe
that RLPGA achieves the best results under all noise ratios
in all tasks. Especially, when the noise ratio is equal to 0.6,
the accuracy of RLPGA is at least 5\% higher than the other
five methods on average.

\begin{figure*}[!h]
	\vskip 0.2in
	\begin{center}
		\centering
		{\includegraphics[scale=0.8]{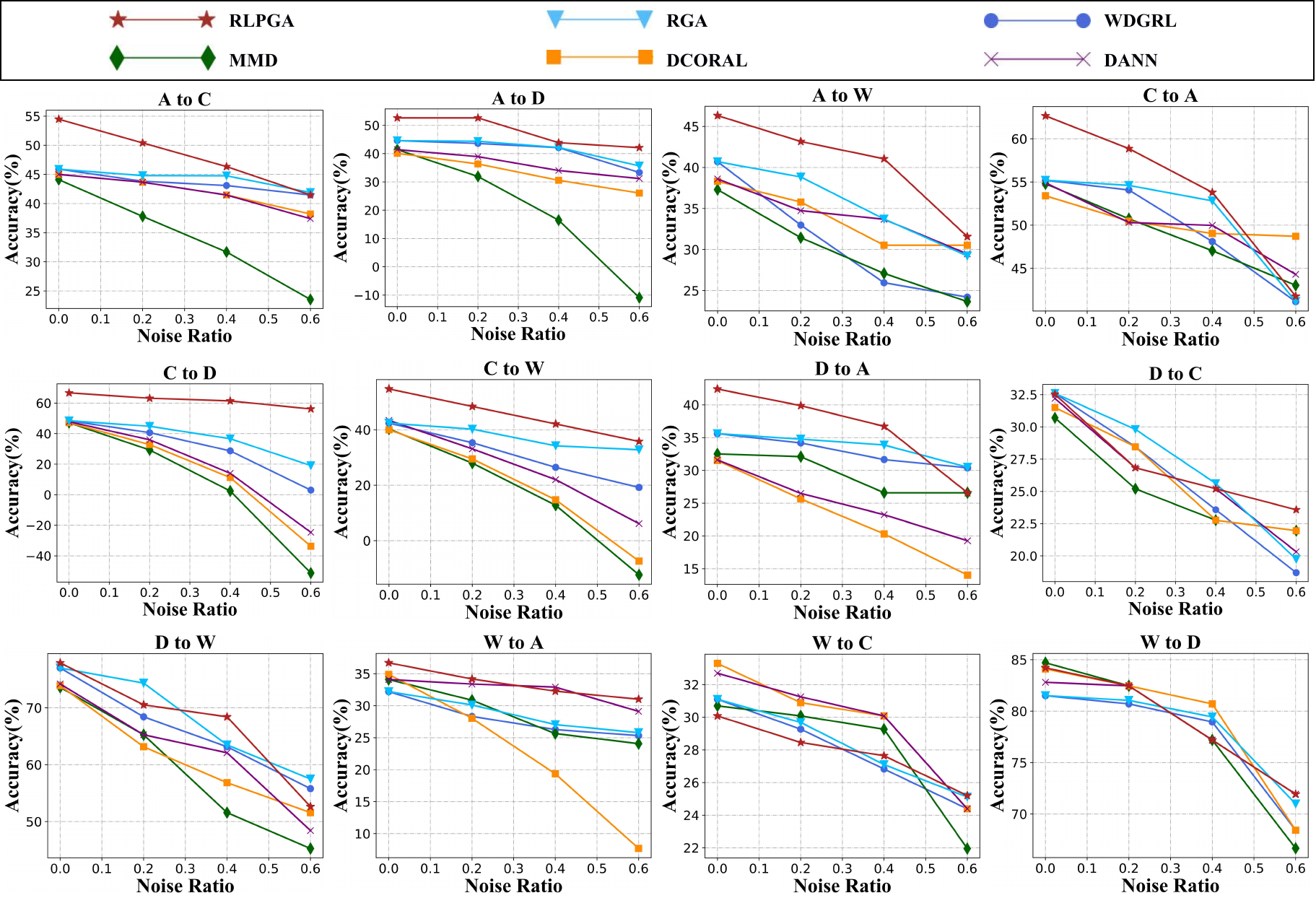}}
		\caption{Robustness evaluation on Office+Caltech10 dataset with SURF features}
		\label{fig:cc}
	\end{center}
	\vskip -0.2in
\end{figure*}

\textit{3) Comparisons on the Office+Caltech10 dataset with SURF features:} From Fig. \ref{fig:cc}, we observe that RLPGA achieves the
best results on 32 of 48 specific transfer tasks. Especially, for
A to D, A to W, C to D, and C to W transfer tasks, RLPGA
achieves the best results on all different noise ratios.

\subsection{Denoising comparisons with random noise} 
\label{sec:dcrncomp}

In order to clarify the performance of our proposed method in different noise circumstances, we conduct extended experiments on Office-Home dataset when the labels of source domain samples are polluted by \textit{random noise}. We set the noise rates in the range of \{0, 0.1, 0.2, 0.3\}, because the difficulty of remaining consistent performance under random noise is much higher than that of keeping robustness under the designed noise, which is based on explicit noise transition matrices of case (1) and (2).

\begin{figure*}[!h]
	\vskip 0.2in
	\begin{center}
		\centering
		{\includegraphics[scale=0.8]{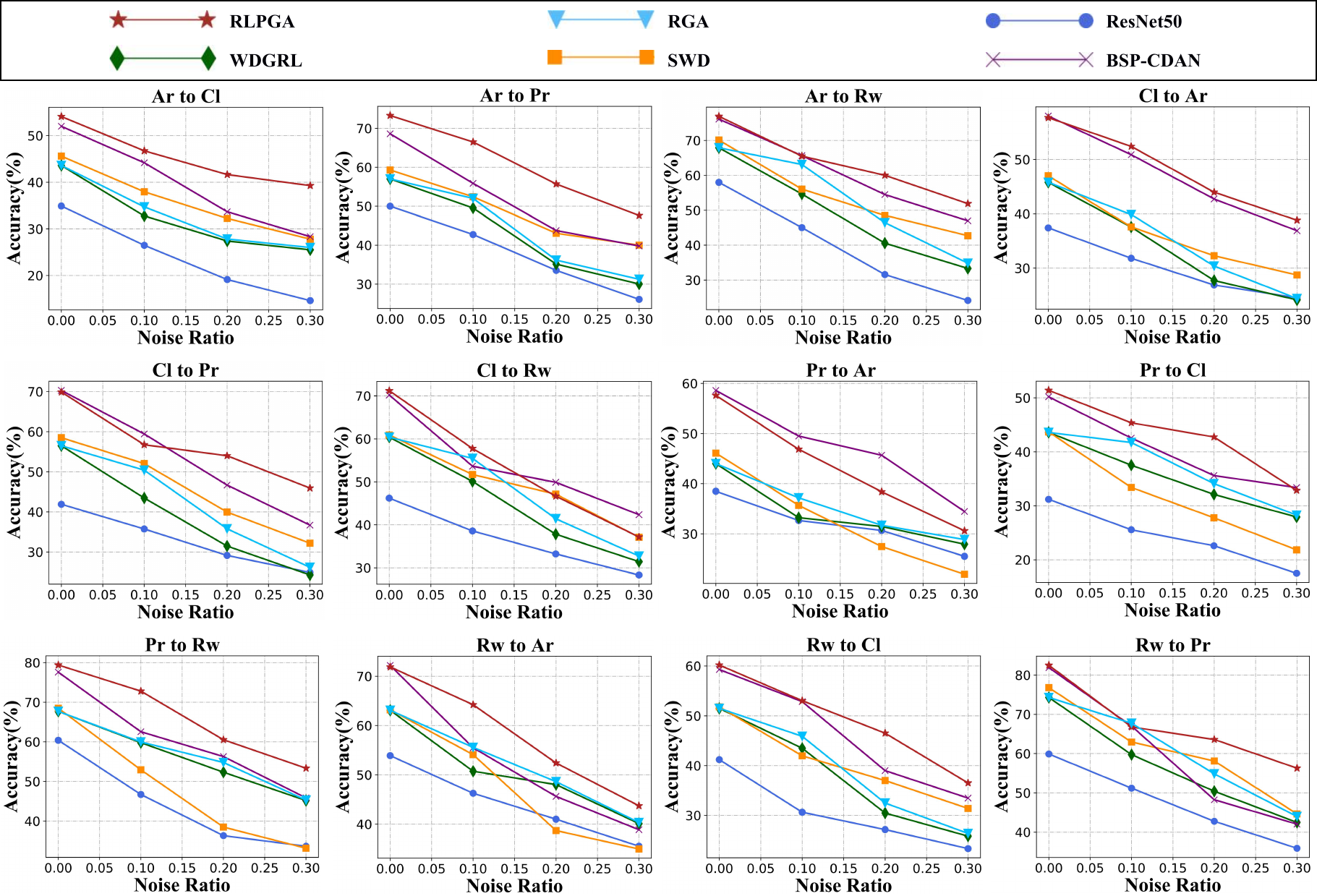}}
		\caption{Robustness evaluation on Office-Home dataset with random noise}
		\label{fig:dcrn}
	\end{center}
	\vskip -0.2in
\end{figure*}

Fig. \ref{fig:dcrn} shows the experimental results on Office-Home dataset with different noise ratios (0, 0.1, 0.2, and 0,3). We compare RLPGA with the ablation model, i.e., RGA, and the other 3 benchmark methods, i.e., BSP+CDAN, SWD, and WDGRL. To understand the robustness of the compared methods, we further perform the backbone method, i.e., ResNet50, and evaluate it within different noise rates. From the figure, we observe that under different noise ratios, RLPGA achieves the best results in most tasks. Especially, when the noise ratio is equal to 0.3, the accuracy of RLPGA is 3.8\% higher than the best benchmark methods on average. We can find that the accuracies of the alternative methods are not very high under the random noise, but compared with other methods, our proposed RLPGA is still able to maintain the robustness to some extent.

\subsection{Detailed influence of hyper-parameters}
\begin{figure*}[!h]
	\vskip 0.1in
	\begin{center}
		\centering{\includegraphics[scale=1]{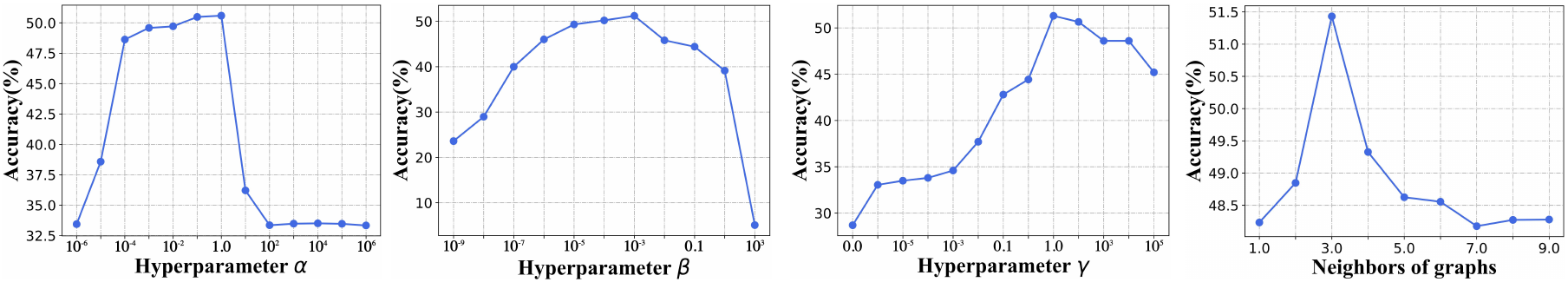}}
		\caption{The influence of parameters }
		\label{pppkey}
	\end{center}
	\vskip -0.25in   	
\end{figure*}

There are four hyper-parameters in our proposed RLPGA including $\alpha$ which is used to balance the impact of the term $Di{s_{pos - neg}}\left( f \right)$, $\beta$ which is used to balance the impact of the term ${W_p}\left( {P_s^{f\left( X \right)},P_t^{f\left( X \right)}} \right)$, $\gamma$ which is used to balance the impact of the term $L_{RIT}$, and the number of neighbor points $k$ when constructing two weight matrices. To understand the influences of the four parameters intuitively, we take some experiments based on the transfer task Pr$\to$Cl of Office-Home dataset. To explore the influence of $\alpha$, we first fix $\beta={10^{-3}}$, $\gamma=1$ and $k=3$ and then select the $\alpha$ from range of $\left\{ {{\rm{1}}{{\rm{0}}^{{\rm{ - 6}}}},...,{\rm{1}}{{\rm{0}}^{{\rm{ - 1}}}}{\rm{,1,1}}{{\rm{0}}^{\rm{1}}},...,{\rm{1}}{{\rm{0}}^{\rm{6}}}} \right\}$. The results are shown in Fig. \ref{pppkey}. We observe that appropriate enhancement of feature discrimination can promote the performance of our proposed method. To explore the influence of $\beta$, we first fix $\alpha=1$, $\gamma=1$ and $k=3$ and then select the $\beta$ from range of $\left\{ {{\rm{1}}{{\rm{0}}^{ - 9}},...,{{10}^3}} \right\}$. From Fig. \ref{pppkey}, we observe that the transferability of learned feature representation is important to the classification task. To explore the influence of $\gamma$, we first fix $\alpha={1}$, $\beta={{10}^3}$, $k=3$, and then select the $\gamma$ from range of $\left\{ {0,{{10}^{ - 5}},...,1,...,{{10}^5}} \right\}$. From Fig. \ref{pppkey}, we observe that the cross entropy loss and $L_{RIT}$ are all important to the classification task. To explore the influence of the number of neighbor points $k$, we first fix $\alpha={1}$, $\beta={{10}^3}$, $\gamma=1$, and then select the $k$ from range of $\left\{ {1,...,9} \right\}$. The results are shown in Fig. \ref{pppkey}, we can see that an appropriate number of neighbor points is important.

\end{document}